\documentclass[journal]{IEEEtranTIE}
\usepackage{graphicx}
\usepackage{cite}
\usepackage{picinpar}
\usepackage{amsmath}
\usepackage{url}
\usepackage{flushend}
\usepackage[latin1]{inputenc}
\usepackage{colortbl}
\usepackage{soul}
\usepackage{multirow}
\usepackage{pifont}
\usepackage{color}
\usepackage{alltt}
\usepackage[hidelinks]{hyperref}
\usepackage{enumerate}
\usepackage{siunitx}
\usepackage{breakurl}
\usepackage{epstopdf}
\usepackage{pbox}

\usepackage{amsfonts,amssymb,amsmath,mathtools,multicol}
\usepackage{physics}

\usepackage{float}
\usepackage{graphicx}
\graphicspath{{images/}}

\usepackage{cite}
\usepackage{hyperref}
\usepackage{subcaption}
\usepackage[dvipsnames]{xcolor}

\usepackage{algorithm}
\usepackage[indLines=true, 
		     noEnd=true, 
		     beginLComment=//~, 
		     endLComment=~, 
		     beginComment=//~]{algpseudocodex}

\usepackage{tabularx}   
\usepackage{booktabs}   

\newcommand{\V}[1]{{\boldsymbol{\mathbf{#1}}}}

\newcommand{\Vdot}[1]{\dot{\V{#1}}}

\newcommand{\R}{\mathbb{R}}
\newcommand{\rarr}{\rightarrow}

\newcommand{\T}{\top}

\newcommand{\Def}{\coloneqq}

\begin{document}
\title{Recasting Classical Motion Planning for Contact-Rich Manipulation}

\author{
	\vskip 1em
	
	Lin Yang, \emph{Student Member, IEEE},
        Huu-Thiet Nguyen,
        Chen Lv, \emph{Senior Member, IEEE},
	\\ and Domenico Campolo$^*$, \emph{Member, IEEE}

	\thanks{
		All authors are with the School of Mechanical and Aerospace Engineering, Nanyang Technological University (NTU), Singapore.
		
		$^*$ Corresponding author: {\tt d.campolo@ntu.edu.sg} 
	}
}

\maketitle
	
\begin{abstract}
In this work, we explore how conventional motion planning algorithms can be reapplied to contact-rich manipulation tasks. Rather than focusing solely on efficiency, we investigate how manipulation aspects can be recast in terms of conventional motion-planning algorithms.
Conventional motion planners, such as Rapidly-Exploring Random Trees (RRT), typically compute collision-free paths in configuration space. However, in many manipulation tasks, contact is either unavoidable or essential for task success, such as for creating space or maintaining physical equilibrium.
As such, we presents Haptic Rapidly-Exploring Random Trees (HapticRRT), a planning algorithm that incorporates a recently proposed optimality measure in the context of \textit{quasi-static} manipulation, based on the (squared) Hessian of manipulation potential. 
The key contributions are $i)$ adapting classical RRT to operate on the quasi-static equilibrium manifold, while deepening the interpretation of haptic obstacles and metrics; $ii)$ discovering multiple manipulation strategies, corresponding to branches of the equilibrium manifold. $iii)$ validating the generality of our method across three diverse manipulation tasks, each requiring only a single manipulation potential expression.
The video can be found at \url{https://youtu.be/R8aBCnCCL40}.
\end{abstract}


\markboth{IEEE TRANSACTIONS ON INDUSTRIAL ELECTRONICS}%
{}

\definecolor{limegreen}{rgb}{0.2, 0.8, 0.2}
\definecolor{forestgreen}{rgb}{0.13, 0.55, 0.13}
\definecolor{greenhtml}{rgb}{0.0, 0.5, 0.0}

\section{Introduction}

\begin{figure}[h!]
    \centering
    \vspace{-6mm}
    \begin{subfigure}[b]{0.45\linewidth}
        \centering
        \includegraphics[width=\linewidth]{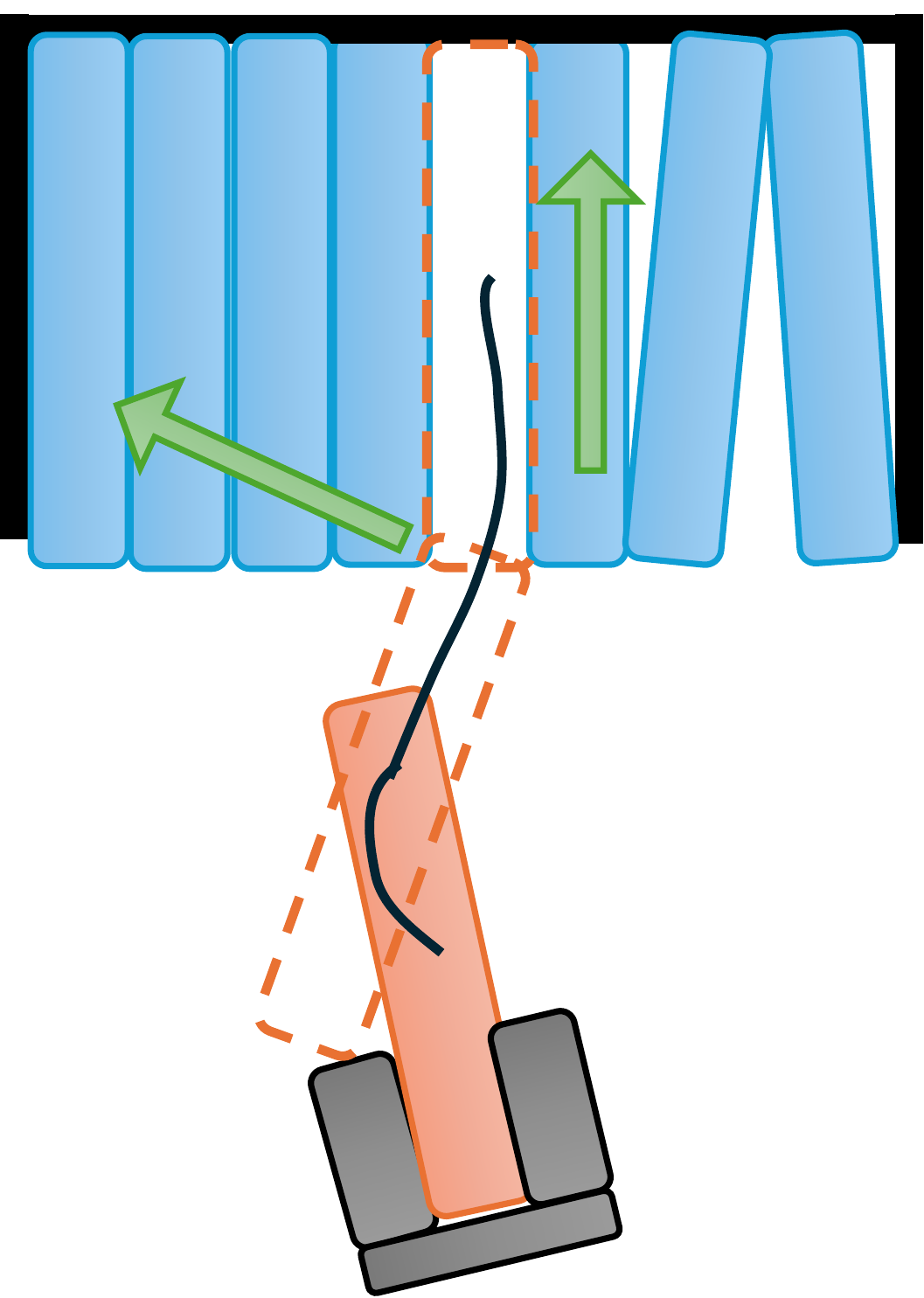}
        \caption{Inserting a book into a crowded shelf by first pushing aside neighboring books to create enough space for a new book.}\label{fig:intro}
    \end{subfigure}
    \hfill
    \begin{subfigure}[b]{0.5\linewidth}
        \centering
        \begin{subfigure}[b]{\linewidth}
            \centering
            \includegraphics[width=\linewidth]{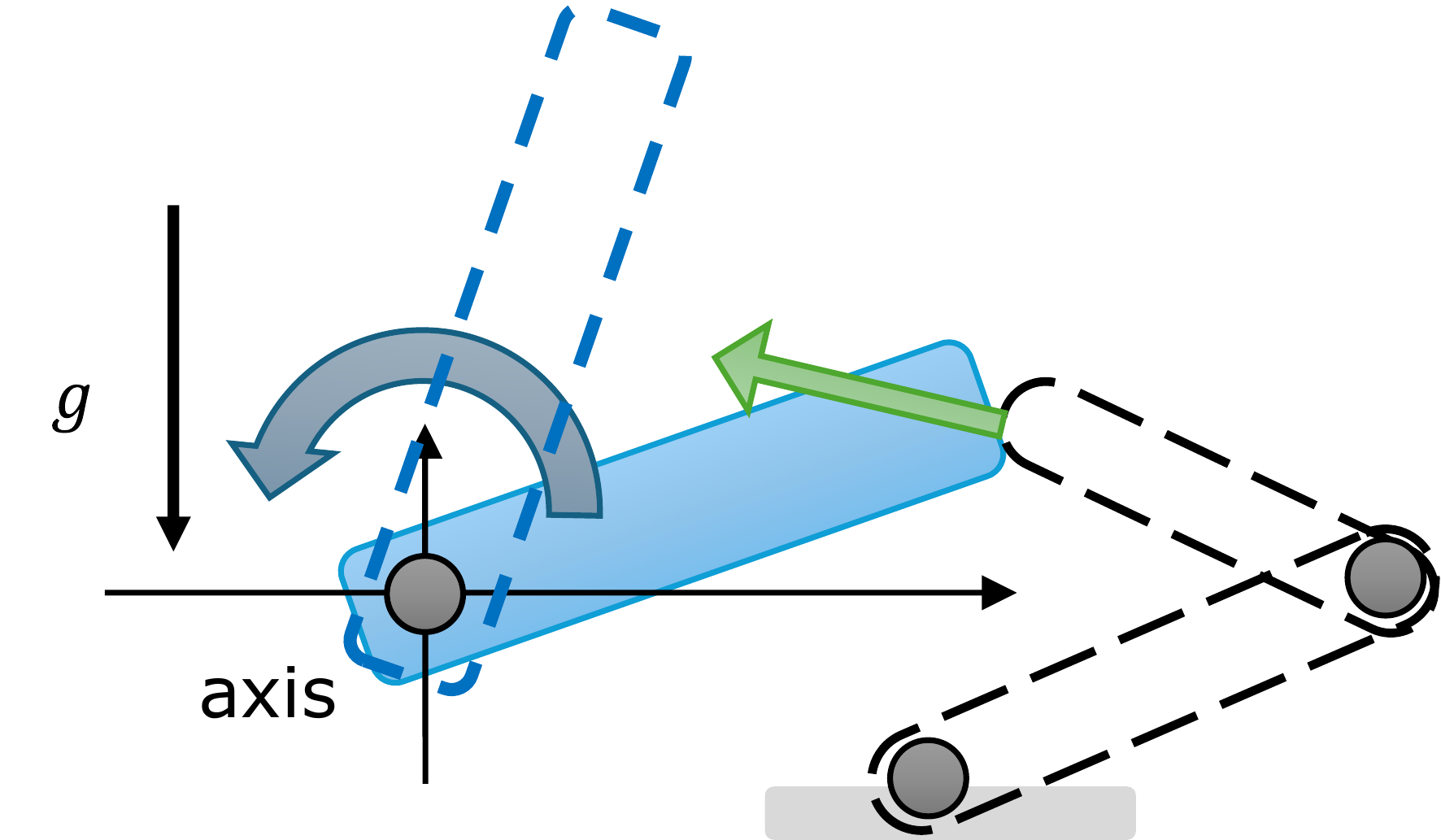}
            \caption{Pushing a hinged pendulum to a desired angle by applying sustained directional force.}\label{fig:intro:slip}
        \end{subfigure}
        \medskip
        \begin{subfigure}[b]{\linewidth}
            \centering
            \includegraphics[width=\linewidth]{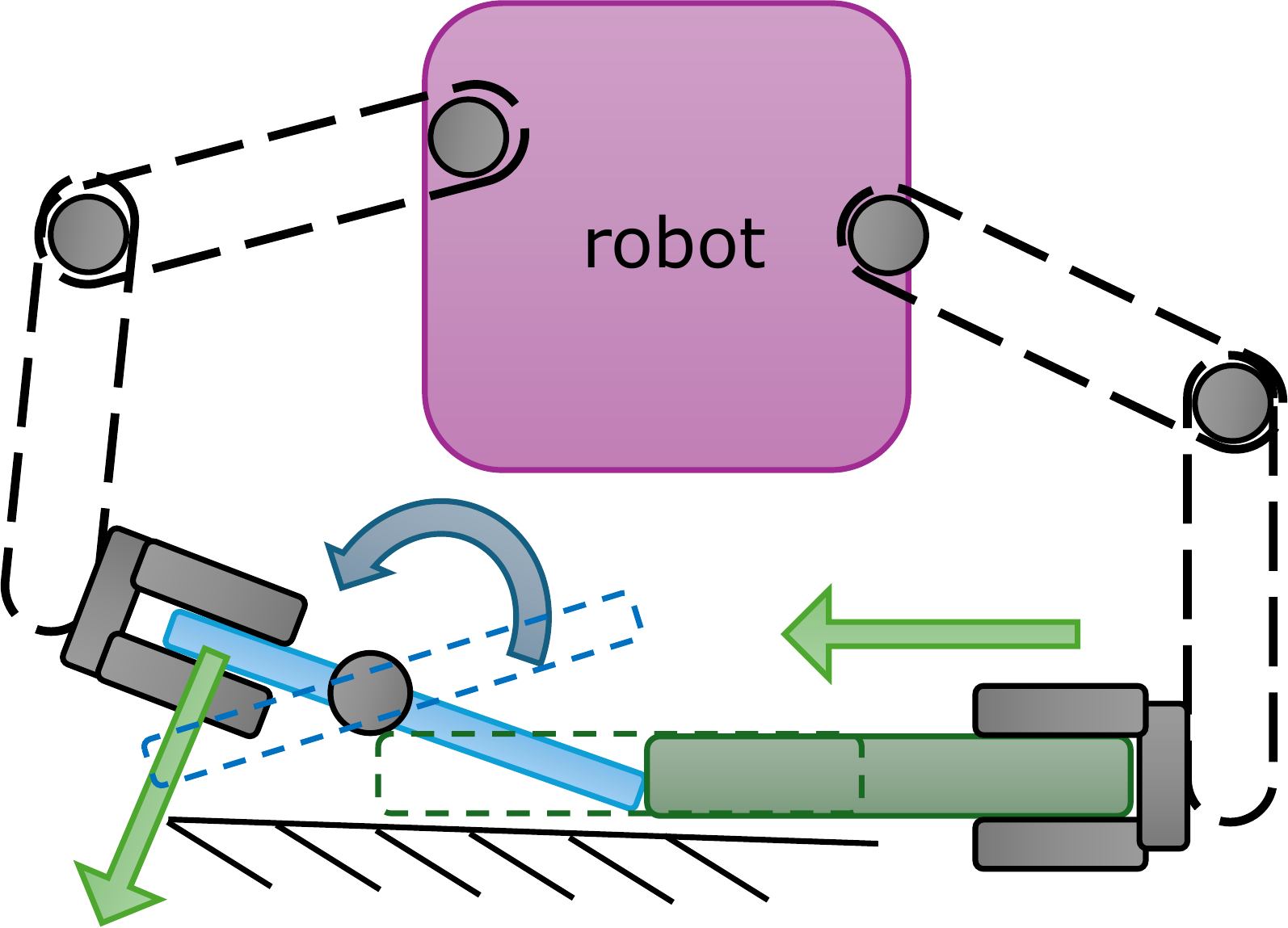}
            \caption{Opening a spring-loaded clip with one arm before inserting an object.}\label{fig:intro:stab}
        \end{subfigure}
    \end{subfigure}
    \caption{Three manipulation tasks require strategic force policy.}
    \label{fig:abc}
\end{figure}

\IEEEPARstart{R}{obotic} manipulation typically involves the robot establishing contact with specific objects. It is essential for the robot to maintain contact with objects to successfully accomplish the tasks \cite{suomalainen2022survey}
. Classical motion planners, such as RRT, sample the configuration space to compute feasible paths while avoiding obstacles. However, in contact-rich manipulation, interactions between the robot and objects are essential for task success.
For example, Fig.\ref{fig:abc} presents three contact-rich manipulation tasks that require purposeful force interaction: (a) inserting a book into a crowded shelf, which involves pushing aside surrounding books before insertion; (b) pushing a hinged pendulum, where the robot must apply directional force to influence a rotating object under gravity; and (c) manipulating a spring-loaded clip, where one arm must apply continuous force to open the clip before the other inserts an object. These tasks demand strategic planning over contact interactions. Traditional planners may easily fail as they do not account for force interactions and the need for controlled contact. This challenge highlights the necessity of a framework that integrates motion and contact interactions while evaluating different manipulation strategies.

Sampling-based methods, including rapidly-exploring random trees (RRT) \cite{lavalle1998rapidly}, have proven to be effective for motion planning \cite{jimenez2024visualizing}. However, their reliance on collision avoidance makes them unsuitable for contact-intensive tasks. To address contact constraints, some approaches formulate the problem within a constraint manifold \cite{kingston2018sampling}, 
leading to methods such as AtlasRRT \cite{jaillet2012path} and IMACS \cite{kingston2019exploring}.
However, these solutions primarily handle geometric constraints and can fail in various scenarios, such as when an object to be inserted is obstructed by other objects. 
Recent work \cite{morgan2022complex} extends planning to both the robot joint space and the object configuration space but does not explicitly capture the force interactions required to rearrange obstructing objects. Other approaches have shown that constructing a spatiotemporal manifold can effectively handle complex geometric constraints \cite{zhou2023spatiotemporal}, but these also neglect force interactions.

A widely adopted approach to incorporating force interactions in manipulation is the quasi-static assumption \cite{whitney1982quasi,ozawa2017grasp,yang2023planning,yang2025planning_book}, which simplifies the problem by focusing on contact forces while neglecting inertial and Coriolis effects.
Recent studies \cite{CAMPOLO2025116003,campolo2023quasi} have demonstrated that quasi-static assumption offers significant theoretical advantages, as it allows force interactions to be modeled as derivable from a smooth potential. This potential unifies robot impedance control and physical contact modeling, enabling manipulation tasks to be framed as an optimization problem based on an intrinsic Riemannian metric (so-called haptic metric), defined as the squared Hessian of the reduced potential \cite{CAMPOLO2025116003}. 
Within this framework, system variables are separated into internal states $\V z$ and control inputs $\V u$, where the control inputs $\V u$ guide the movement of indirectly controllable objects $\V z$ along an implicitly defined equilibrium manifold ($\mathcal M^{eq}$).
Our earlier work \cite{yang2025planning_book} showed how to navigate on $\mathcal M^{eq}$ and compute optimal control policies, but a systematic exploration of implicit manifold and clear visualizations of key concepts were not provided.

While quasi-static manipulation provides a structured approach to analyzing contact-rich tasks, determining a control policy for mechanical systems remains an open challenge. Traditional quasi-static methods often require extensive manually defined contact phases \cite{whitney1982quasi,salem2020robotic,ozawa2017grasp}, limiting their flexibility. Similarly, learning from demonstration (LfD) approaches \cite{wang2024cooperative, chen2024robust} rely on human-provided trajectories and encode task knowledge through manual demonstrations. 
On the other hand, Reinforcement learning (RL) has been explored as an alternative \cite{elguea2023review}, but it typically relies on task-specific reward functions, suffers from long training times, and faces the curse of dimensionality \cite{bing2022solving}. Conversely, classical planning algorithms (e.g., RRT) are computationally efficient in high-dimensional spaces, though they are not directly applicable to contact-rich tasks.
Motivated by these challenges, our key contributions are as follows:  

\begin{enumerate}[1)]
  \item \textbf{Sampling-based planning for contact-rich manipulation:} We adapt the classical RRT planner to a quasi-static formulation, introducing HapticRRT, a method that plans over an implicit equilibrium manifold $\mathcal{M}^{eq}$ and incorporates visual tools to reveal how haptic metrics and obstacles emerge within this framework, providing intuitive insights into contact-rich planning.
  \item \textbf{Exploration of multiple manifold branches:} We introduce and interpret the concept of multiple branches in $\mathcal{M}^{eq}$, highlighting their practical significance for success of manipulation tasks.
  \item \textbf{Validation across diverse manipulation tasks:} We evaluate HapticRRT on three representative contact-rich scenarios, demonstrating that HapticRRT discovers strategic manipulation behaviors in each case.
\end{enumerate}

To demonstrate the generality and significance of our approach, we evaluate HapticRRT on three manipulation tasks that represent different aspects of contact-rich planning.
First, in a \textit{pendulum manipulation} task, the robot must strategically apply force on an underactuated pendulum. Unlike the classical inverted pendulum \cite{irfan2024control}, this task more closely resembles door handles \cite{shaikh2023door}.
Second, in a \textit{spring-loaded clip manipulation} task, rather than using dexterous hands to squeeze the clip \cite{kim2021integrated}, we demonstrate non-prehensile manipulation using a standard two-finger gripper.
Third, in a \textit{crowded book insertion} scenario, prior methods \cite{nakajima2011study, sygo2023multi} often rely on carefully designed, task-specific hierarchical policies to rearrange clutter before insertion.
In all three tasks, HapticRRT autonomously discovers strategic manipulation policies and identifies branches of the manifold, demonstrating its ability for generalized contact-rich planning.

\vspace{-2mm}
\section{Manipulation Planning on the Implicit Equilibrium Manifold}\label{sec: equilibrium mfd}

Building upon our previous work \cite{yang2025planning_book}, we briefly introduce the key concepts of our framework, including the equilibrium manifold, haptic metric, and haptic obstacle, to ensure a self-contained presentation. The novel contributions in this paper lie in the introduction of multiple equilibrium branches, which we formally define in Sec. \ref{sec: multiple branches} and apply classical motion planner RRT into our framework, detailed in Sec. \ref{sec: hapticRRT}. Furthermore, 3 separate representative tasks and their manifold are presented in Sec. \ref{sec: IP}, \ref{sec:book}, \ref{sec:clip}.

\vspace{-2mm}
\subsection{Quasi-Static Mechanical Manipulation System}

Under quasi-static assumption, we describe the environment (objects) and robots as an \textit{interconnected system} $\mathcal{Z} \times \mathcal{U}$ \cite{CAMPOLO2025116003,campolo2023quasi}, where $\V z\in\mathcal{Z}\subset\R^{N}$ represents the \textit{internal state} (also referred to as indirectly controllable objects) and $\V u\in\mathcal{U}\subset\R^{K}$ is the \textit{control} of the robot (which can be interpreted as the desired pose in impedance control).
The configuration of the system is determined solely by its manipulation potential $W(\V z, \V u)$, such as elastic and gravitational energies.
Define manipulation potential as a smooth field on the space $W: \mathcal{Z}\times\mathcal{U} \rarr \R$. Equilibria $\V z^*$ are found from
\begin{equation}
\partial_{\V z}W(\V z^*, \V u) = \V 0 \in\R^{N}.
\label{eq: W_z}
\end{equation}
We define  $\partial_{\V q} W \equiv [\partial_{q_1} W, \ldots, \partial_{q_a} W]^T$, where $\partial_{\V q} = [\partial_{q_1}, \ldots, \partial_{q_a}]^T$. Meanwhile, define the shorthand notation $\partial^2_\V {zz} \equiv \partial_{\V z} \partial_{\V z}^T$ for Hessians and mixed-derivative operators. 
Here, $\partial_{\V z}$ denotes the gradient with respect to $\V z$, which means internal forces acting on objects $\V z$. Under quasi-static assumption, the total force acting on the objects should be zero. We describe the interplays of objects and a robot, i.e., $\V f_\text{ctrl} =  -\partial_\V{u}W$ the so-called \textit{control forces} \cite{CAMPOLO2025116003}.

A point is stable when its Hessian is positive definite, i.e., $\partial^2_\V{zz}W|_{\ast} \succ 0$. Assuming the Hessian $\partial^2_\V{zz}W\in\R^{N\times N}$ is of full rank when $\partial_{\V z}W(\V z^*, \V u) = \V 0$, via the \textit{implicit function theorem} \cite{spivak2018calculus}, the set
\begin{equation}
\mathcal{M}^{eq}\Def \{(\V z,\V u)\in\mathcal{Z}\times\mathcal{U} | \partial_{\V z}W(\V z,\V u)=\V 0\}
\end{equation}
is a smooth embedded submanifold in the ambient space $(\mathcal{Z} \times \mathcal{U})$. We refer to $\mathcal{M}^{eq}$ as the \textit{equilibrium manifold} (EM) of the system. The state transitions are purely controlled by  $\V u$. Thus, to guarantee the stability, the control should avoid getting close to singularities. Therefore, define haptic obstacle as 
\begin{equation}\label{eq: Haptic Obstacle}
\rm{det} ( \partial^2_\V {zz} W (\V z, \V u)) > \lambda > 0
\end{equation}
where $\lambda>0$ is a threshold based on stiffness.

\vspace{-2mm}
\subsection{Multiple Branches of Manifold}\label{sec: multiple branches}

\begin{figure}[!ht]
\centering
\includegraphics[width=0.35\textwidth]{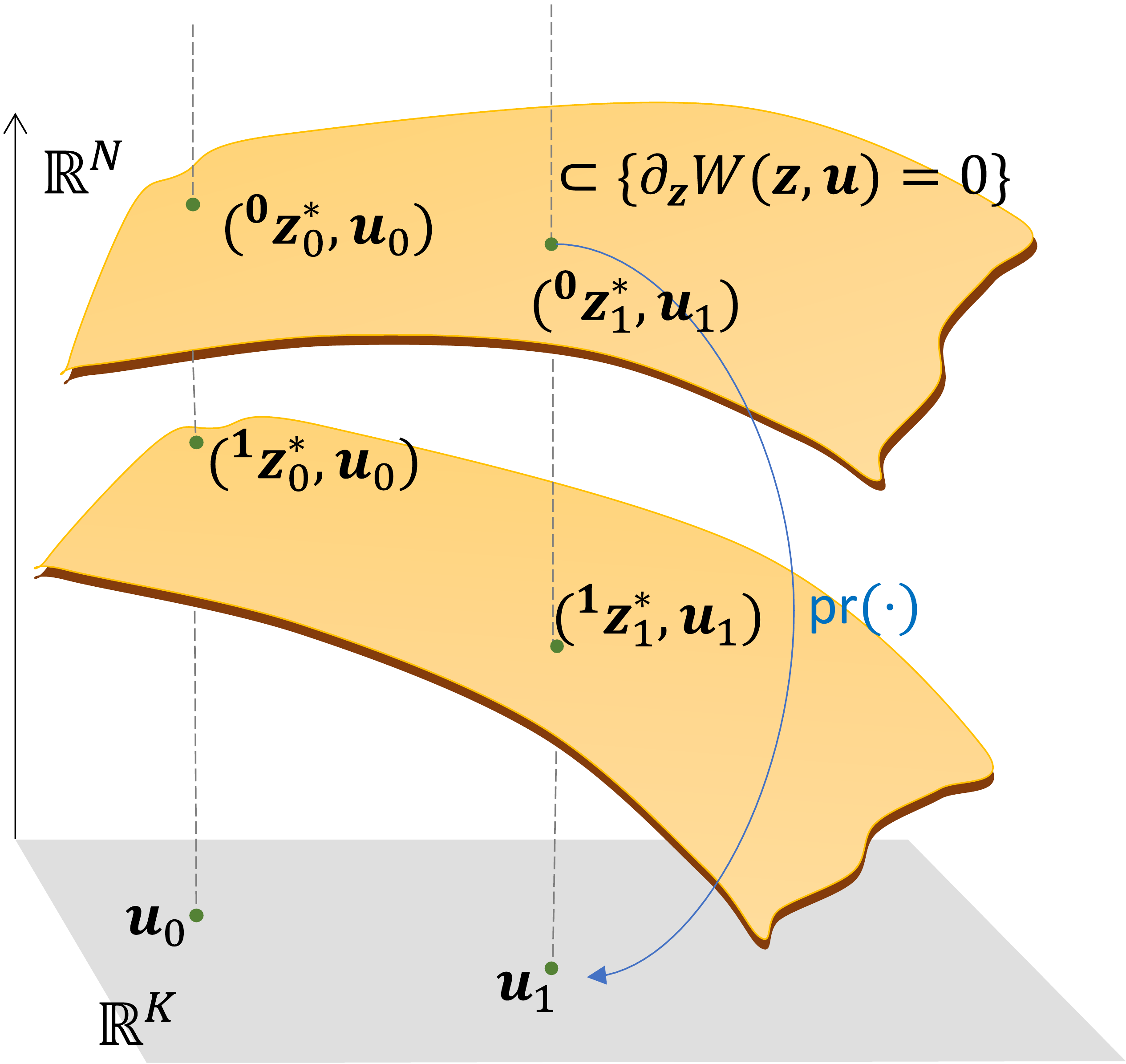}
\caption{Configuration space $(\mathcal Z \times \mathcal U)$ and multiple branches of equilibrium manifolds. For same control $\V u$, there could exist several internal state $^m\V z^*_i$.}
\label{fig: ambient space}
\end{figure}

Note, for quasi-static manipulations, solutions are often \textit{multi-valued}, e.g., manipulating an object with two hands, there may exist multiple stable configurations for the same grasping pose. Consequently, the equilibrium manifold $\mathcal{M}^{eq}$ could contain multiple branches, as depicted in Fig. \ref{fig: ambient space}. Additionally, each stable solution $^m\V z^*_i$, with $m \geq 1$ indicating multiplicity of equilibria, can only be identified  after specifying the input $\V u_i$, leading to a natural projection,
\begin{equation}
\rm{pr} : (^m\V z^*_i, \V u_i) \mapsto \V u_i
\label{eq:projection}
\end{equation}
In practical terms, the existence of multiple branches means that same control policies can lead to distinct object states, depending on the historical control policy.

\vspace{-5mm}
\subsection{Haptic Metric and Haptic Distance}\label{sec: haptic metric}

The notion of closeness between states is determined by a distance function. Following \cite{CAMPOLO2025116003,campolo2023quasi}, we defined the Riemannian metric of the control space $\mathcal{U}$, where the squared Hessian $\V G^2_m(\V z^*(\V u), \V u)$ is called the \textit{haptic metric}, which offers a more general measure of interaction.
\begin{equation}\label{eq: control Hessian}
\V G_m(\V z^*(\V u), \V u) \Def \partial^2_\V{uu}W - \partial^2_\V{uz}W (\partial^2_{\V{zz}}W)^{-1} \partial^2_\V{zu}W,
\end{equation}
which is computed as the Schur complement of the Hessian of the potential function $W(^m \V z^*, \V u)$, evaluated at equilibrium (i.e., $^m\V z^*(\V u)$ s.t. $\partial_\V{z}W(^m\V z^*, \V u)=\V 0$).

For any control policy $\V u(s) : [0,1]\rarr\R^K$ connecting two points in the control space, haptic distance $S$ between any two points $\V u(0)$ to $\V u(1)$ is defined as,
\begin{equation}\label{eq: geodesic U}
S[\V u] = \int_{0}^{1} \sqrt{\dot{\V {u}}^{T} \V G^{2}_m(\V z^*(\V u), \V u) \dot{\V {u}}} \; ds
\end{equation}
The greater force exerted by robot, the larger the value of $S$.

\section{HapticRRT} \label{sec: hapticRRT}

We have introduced the basic framework, and the objective is to manipulate objects $\V z$ to a desired position based on the task requirements. However, since $\V z$ is implicitly defined, the exact value of $\V z^*(\V u)$ remains unknown.
In this section, we present how classical sampling-based motion planners, RRT \cite{lavalle1998rapidly}, can be integrated into our framework. By leveraging the tree structure of RRT, we explore the implicit equilibrium manifold until a feasible path connecting the initial state to the desired state is found.

\subsection{Sampling in Control Space}

Following the classical RRT approach, we assume that a tree $\mathcal{T}$ is being incrementally constructed. At each iteration, a random node is selected. However, instead of sampling from the entire configuration space, we restrict our selection to the control space $\mathcal{U}$, choosing a random control input $\V u_\text{rand}$.

Next, we determine the nearest node in the control space, denoted as $\V u_\text{near}$, and pair it with its corresponding state to form $(\V z_\text{near}, \V u_\text{near})$. 
Unlike standard RRT, this nearest neighbor selection considers both the Mahalanobis distance and the manipulation potential $W(\V z, \V u)$. 
While proximity in configuration space remains important, the algorithm is biased toward nodes with lower potential.
This reflects a trade-off: some contact is required to accomplish manipulation tasks, but excessive contact may indicate that the robot is stuck.
Therefore, the revised distance incorporates both geometric proximity and energetic feasibility.
The geometric term is represented by the Mahalanobis distance $\norm{\V u - \V u_\text{rand} }_\Sigma$, and the energetic term by the manipulation potential $W^\beta(\V z, \V u)$, where $\beta$ is a tunable parameter.
This is implemented in Line 3 of Alg. \ref{alg: SAMPLE}.

Importantly, we consider only nodes where the $\textsc{DeadEnd}$ flag is set to $\texttt{False}$, ensuring that the node remains valid for further expansion. The $\textsc{DeadEnd}$ label indicates whether a state encounters a haptic obstacle (as defined in Eq. \ref{eq: Haptic Obstacle}); only states that do not face haptic obstacle are eligible for tree growth.

In classical RRT, expansion typically proceeds by moving a fixed step toward $\V u_\text{rand}$. However, in our framework, we must adhere to the quasi-static assumption, ensuring that the system remains on the equilibrium manifold. Direct expansion may disrupt continuity or lead to unstable configurations. Therefore, instead of taking a discrete step, we slowly move toward $\V u_\text{rand}$ to maintain stability.

\begin{algorithm}[ht]
\caption{Sample a direction in control space}\label{alg: SAMPLE}
\begin{algorithmic}[1]
\Procedure{SAMPLE}{$\mathcal{U},\mathcal{T}$}
\State $\V u_\text{rand} \gets$ randomly select from $\{\mathcal{U}\}$
\State $\V u_\text{near} \gets \underset{\V u}{\arg \min } \ W^{\beta}(\V z,\V u)\norm{\V u - \V u_\text{rand} }_\Sigma, \; (\V z, \V u) \in \mathcal{T}, \ \textsc{DeadEnd} = \texttt{False}$
\State $\Vdot u = (\V u_\text{rand} - \V u_\text{near})/\norm{\V u_\text{rand} - \V u_\text{near}}_2$ 
\State\Return $ \Vdot u, (\V z^*_\text{near}, \V u_\text{near})$
\EndProcedure
\end{algorithmic}
\end{algorithm}

\subsection{Extending via Adaptive ODE}

To move a node along $\mathcal{M}^{eq}$, we follow the method as in our previous work \cite{yang2025planning_book}, which employs an adaptive Ordinary Differential Equation (ODE) approach:

\begin{equation}
\Vdot z = -(\partial^2_\V{zz}W)^{-1}\partial^2_\V{uz}W \Vdot u - \eta(\partial^2_\V{zz}W)^{-1} \partial_\V z W \label{eq:adaptive ODE}
\end{equation}

\begin{figure}[!ht]
\centering
\includegraphics[width=0.5\textwidth]{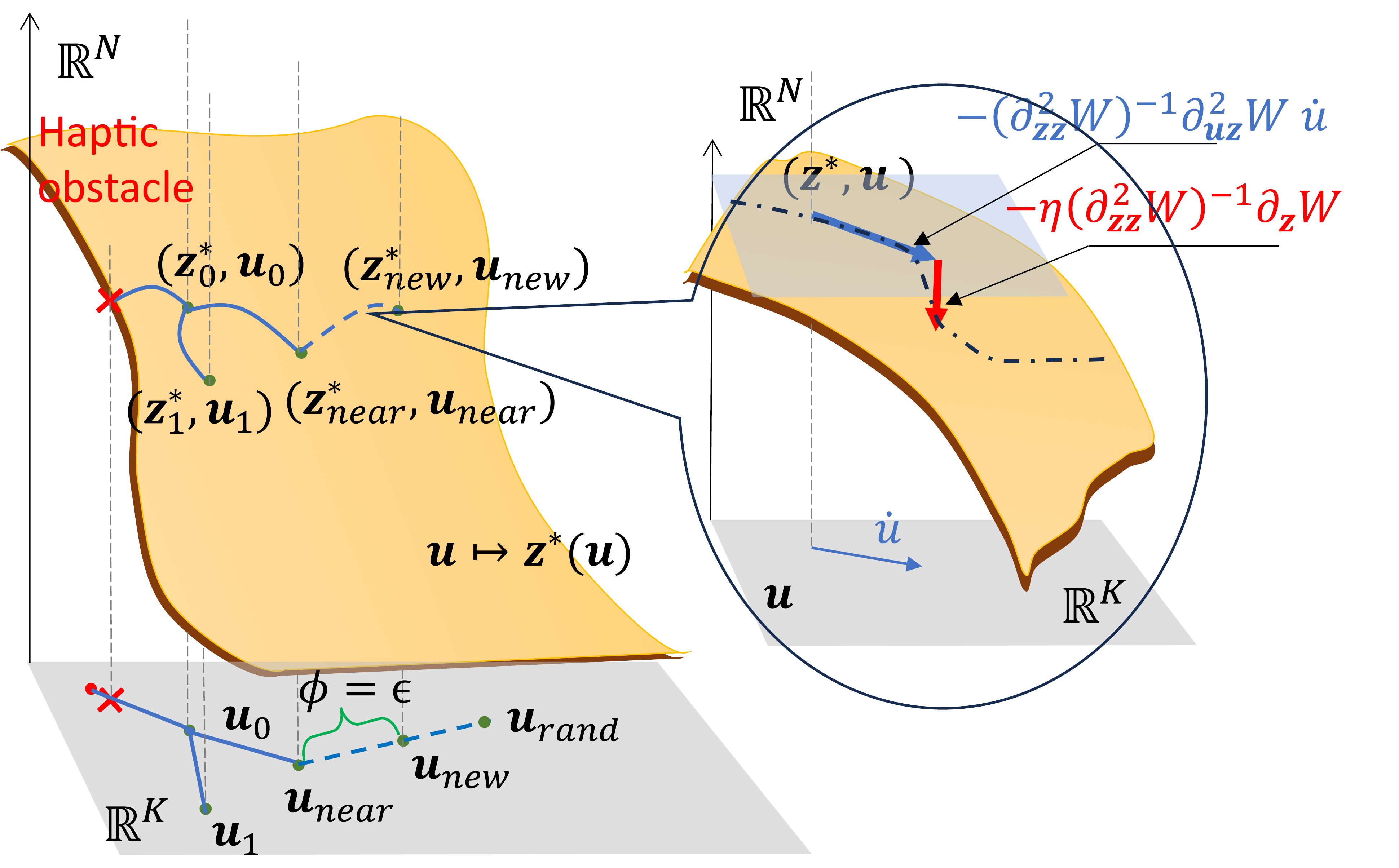}
\caption{Right: The adaptive ODE enables nodes to move along $\mathcal{M}^{eq}$. Left: HapticRRT explores $\mathcal{M}^{eq}$ while ensuring that nodes remain on the manifold until either the haptic distance value reaches $\epsilon$ or the ODE is terminated by haptic obstacle.}
\label{fig: adaptive ODE}
\end{figure}

Eq. \ref{eq:adaptive ODE} consists of two key terms, the former  (depicted as a blue arrow in Fig. \ref{fig: adaptive ODE}) captures the linear relationship between the infinitesimal changes in $\V z$ and $\V u$. The later (represented by the red arrow in Fig. \ref{fig: adaptive ODE}) corresponds to Newton-Raphson infinitesimal adjustments, ensuring that the system remains on the equilibrium manifold. Since holding $\V u$ constant leads to out-of-equilibrium dynamics, this correction term is necessary. The parameter $\eta$ represents the step size.

With this approach, we can track the evolution $t \to \V z(t) \in \R^N$ as the control parameters evolve as $t \to \V u(t) \in \R^K$ by numerically solving the adaptive ODE. This ensures that the tree structure is extended while remaining on EM.

Moreover, similar to RRT strategy of extending the tree by a fixed distance $\epsilon$, we also extend our tree for a predetermined haptic distance.
Within this framework, a functional value of haptic distance $\phi$, as defined in Eq. \ref{eq: geodesic U}, is computed using the ODE, incorporating the haptic metric. Consequently, the ODE governing the entire system can be expressed as follows:

\begin{subequations}
\begin{align}
\frac{d}{dt}
\begin{bmatrix}
	\V z  \\
	  \V u   \\
        \phi
\end{bmatrix} 
&=
\begin{bmatrix}
	-(\partial^2_\V{zz}W)^{-1}\partial^2_\V{uz}W \Vdot u - \eta(\partial^2_\V{zz}W)^{-1} \partial_\V z W  \\
	   (\V u_\text{rand} - \V u_\text{near})/\norm{\V u_\text{rand} - \V u_\text{near}}_2   \\
        \sqrt{\Vdot u^T \V G_{m}^{2}(u) \Vdot u}
\end{bmatrix}  
\\
\begin{bmatrix}
	\V z(0)  \\
	  \V u(0)   \\
        \phi(0)
\end{bmatrix}
&=
\begin{bmatrix}
	\V z^*_\text{near}  \\
	  \V u_\text{near}    \\
        0
\end{bmatrix}
\end{align}
\label{eq:whole ODE}
\end{subequations}

One termination condition occurs when $\phi(t) \leq \epsilon$, at which point we return a new node $(\V z_\text{new}, \V u_\text{new})$ and set \textsc{DeadEnd} = $\texttt{False}$.
A false \textsc{DeadEnd} flag indicates that the node is a valid expansion point for future tree growth. Conversely, if the node encounters a haptic obstacle (as defined in Eq. \ref{eq: Haptic Obstacle}), tree expansion is also terminated.
The \textsc{Extend} function is formally defined in Alg. \ref{alg: local planning}.

\begin{algorithm}[ht]
\caption{Extend on equilibrium manifold}\label{alg: local planning}
\begin{algorithmic}[1]
\Procedure{Extend}{$(\V z_\text{near}, \V u_\text{near}), \Vdot u, \epsilon$}
\State $\V z(t), \V u(t), \V \phi(t) \gets$ solve ODE via Eq. \ref{eq:whole ODE}
\If{$\V \phi(t) > \epsilon$}
	\State Stop, $\textsc{DeadEnd}\gets$ \texttt{False}
\EndIf
\If{$\rm{det} ( \partial_{zz} W (\V z(t), \V u(t))) > \lambda$}
	\State Stop, $\textsc{DeadEnd}\gets$ \texttt{True}
\EndIf
\State\Return $(\V z^*_{new}, \V u_{new}) = (\V z(t), \V u(t)), \V \phi(t)$
\EndProcedure
\end{algorithmic}
\end{algorithm}

\subsection{Overall Algorithm}

Alg. \ref{alg: HapticRRT} presents our final planning framework.
We begin by initializing a stable node $(\V z^*_\text{start},\V u_\text{start})$ on EM, ensuring that the stability condition (Eq. \ref{eq: Haptic Obstacle}) holds. Subsequently, the function \textsc{Sample} returns both a direction and a candidate node for growth, while the function \textsc{Extend} generates a new node on EM. Finally, the new node and its corresponding edge are added to the tree, along with its \textsc{DeadEnd} label to indicate whether further expansion is possible.
The conceptual framework of HapticRRT is illustrated in Fig. \ref{fig: adaptive ODE}.

\begin{algorithm}[!ht]
\caption{HapticRRT}\label{alg: HapticRRT}
\begin{algorithmic}[1]
\Require$(\V z^*_\text{start},\V u_\text{start})\in\mathcal{M}^{eq}$ the starting point on the equilibrium manifold, $\epsilon$ the geodesic size and $N$ the maximum number of attempts.
\Ensure A search tree $\mathcal{T}=(V,E)$.
\State$V\gets\{(\V z_\text{start}, \V u_\text{start})\}$; $E\gets\emptyset$
\For{$n=1,\dots,N$}
	\State$\Vdot u, (\V z^*_\text{near}, \V u_\text{near})\gets$ \Call{Sample}{$\mathcal{U},\mathcal{T}$}
	\State$(\V z^*_\text{new}, \V u_\text{new}), \textsc{DeadEnd}\gets$ \Call{Extend}{$\Vdot u, (\V z^*_\text{near}, \V u_\text{near}), \epsilon$}
 \State $V \gets V \cup \{(\V z^*_\text{new}, \V u_\text{new}), \textsc{DeadEnd}\}$; $E \gets E \cup \{(\V z^*_\text{near}, \V z^*_\text{new}), (\V u_\text{near}, \V u_\text{new})\}$
\EndFor
\State\Return$\mathcal{T}=(V,E)$
\end{algorithmic}
\end{algorithm}

\section{Manipulation of a Pendulum} \label{sec: IP}
In this section, we present a manipulation task involving a rectangular pendulum and a robot. Our approach employs a robot to interact with and manipulate the pendulum, where the motion of the pendulum is driven by the interaction between the robot and the pendulum \cite{CAMPOLO2025116003}.
To model this task, we follow the same mathematical tool as in our previous work \cite{yang2025planning_book,yang2025energy}.


\subsection{Superellipses and Contact Stiffness}
To apply our framework, we require only a \textit{differentiable} manipulation potential. One way to obtain this is by modeling the system using superquadrics (SQ), which, in 2D, are referred to as superellipses \cite{jaklic2000segmentation}. In the following, we introduce key components of our modeling approach.


\subsubsection{Superellipses}
As the shape of the pendulum is rectangular, we model it by a SQ which is implicitly defined by the equation:
\begin{align}
    \left(\frac{x}{a_1}\right)^{\frac{2}{\varepsilon}} + \left(\frac{y}{a_2}\right)^{\frac{2}{\varepsilon}} = 1
    \label{eq:SQ}
\end{align}
where $\varepsilon$ determines the shape of SQ, and $a_1, a_2$ define its size.
To facilitate contact modeling, we rewrite Eq. \ref{eq:SQ} as an \textbf{inside-outside} function $F(x,y)$, given by:
\begin{align}
    F(x,y) &= \left(\frac{x}{a_1}\right)^{\frac{2}{\varepsilon}} + \left(\frac{y}{a_2}\right)^{\frac{2}{\varepsilon}} - 1
    \label{eq:io}
\end{align}
which possesses a useful property. For any given point $(x_0,y_0)$, Eq. \ref{eq:io} determines its relation to SQ: outside if $F(x_0,y_0) > 0$, inside if ($F(x_0,y_0) < 0$), and on the surface if $F(x_0,y_0) = 0$.

\subsubsection{Contact stiffness}

The inside-outside function $F(x,y)$ from Eq. \ref{eq:io} can be leveraged to model contact interaction. To capture contact behavior, we define a nonlinear stiffness function $k(d)$, which decides the contact force:


\begin{equation}\label{eq: stiffness}
k(d) = k_\text{min} + \frac{1-\tanh(d/d_0)}{2} k_\text{max}.
\end{equation}
where $d_0$ is a constant that decides the steepness of the stiffness curve, ensuring a smooth transition between the contact and non-contact states. The parameters $k_\text{max}$ and $k_\text{min}$ represent the maximum and minimum stiffness values, respectively, with $k_\text{max} \gg k_\text{min}$. The independent variable $d$ is computed from $F(x,y)$, expressed in SQ frame. Due to the properties of the inside-outside function:
When the point is outside SQ (non-contact region, $F(x,y) > 0$), the stiffness remains at its minimum value $k_\text{min}$.
When the point is inside SQ (contact region, $F(x,y) < 0$), the stiffness increases, governed by $k(d)$, to reflect contact interaction.

\subsection{Pendulum Modeling}

\begin{figure}[!h]  
\centering
  \begin{subfigure}{0.2\textwidth}
    \includegraphics[width=\linewidth]{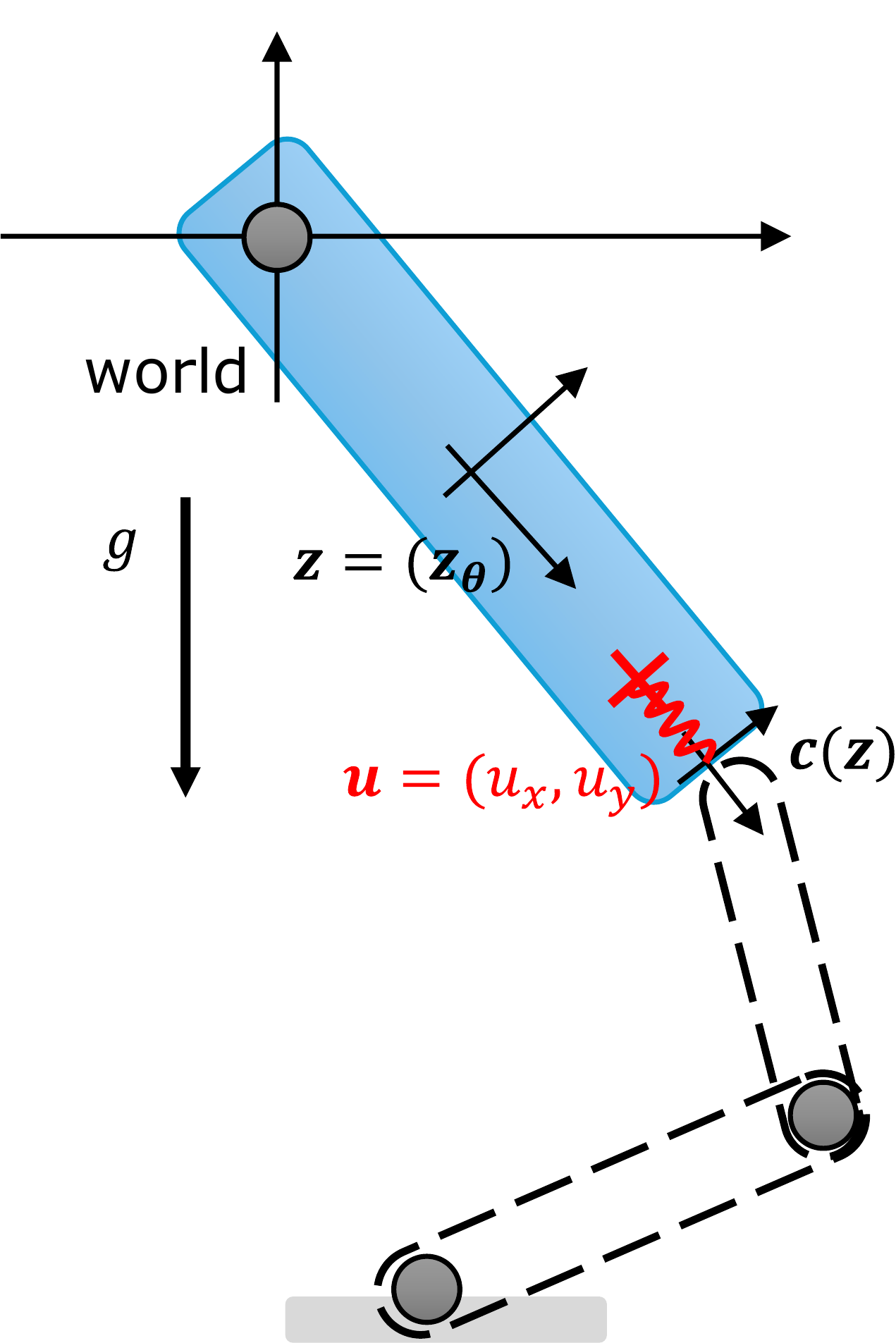}
    \caption[]
    {\small System modeling.}
    \label{fig: inverted pendulum}
  \end{subfigure}%
  \hspace{2mm}  
  \begin{subfigure}{0.22\textwidth}
    \includegraphics[width=\linewidth]{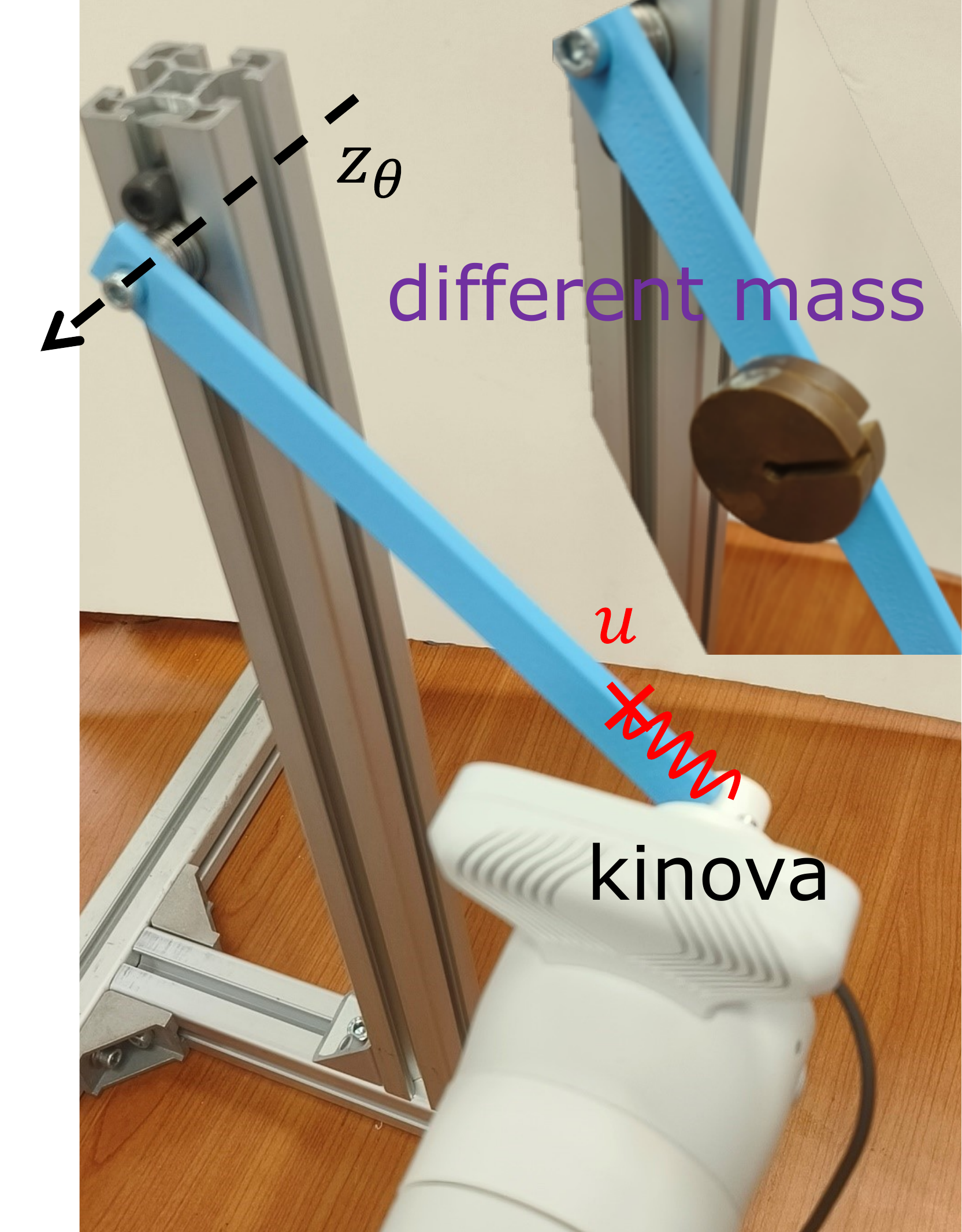}
    \caption[] 
    {\small Real world setup: pendulum with different masses.}
    \label{fig: inverted pendulum exp}
  \end{subfigure}%
\caption{Manipulating a hinged pendulum with varying masses via sustained directional force.} 
\label{fig:2DTU}
\end{figure}

The system consists of a pendulum and a robot in a 2D plane. The pendulum is hinged at one end to the origin with length $L_{0}$, and a body frame is attached at its center of mass (CoM) with mass $m$. As illustrated in Fig. \ref{fig: inverted pendulum}, the system's internal state variable is the pendulum angle, defined as $\V z = z_\theta \in S^1$.
A 2D point robot interacts with the tip of the pendulum, applying forces to manipulate its motion. The robot is denoted by $\V u = [u_x, u_y]^\T \in \R^2$. Through this interaction, the robot indirectly controls the pendulum.
The manipulation potential of the system is defined as:
\begin{align}
W(\V z, \V u) = & W_\text{grav}(\V z) +W_\text{contact} (\V z, \V u),\nonumber \\
= & \frac{1}{2} mg L_{0} \sin z_\theta+ \frac{1}{2} k \bigg((u_{x}-L_{0} \cos z_\theta)^{2}  \nonumber \\
&+ (u_{y}-L_{0} \sin z_\theta)^{2} \bigg),
\end{align}
where $W_\text{grav}(\V z)$ represents the gravitational potential of the pendulum, $W_\text{contact}(\V z, \V u)$ captures the interaction energy between the pendulum and the robot.
Other derivative terms can be computed analytically.

\subsection{HapticRRT for Pendulum Manipulation}

In previous work, Campolo et al. \cite{campolo2023quasi} computed EM for this system, demonstrating that the manipulation of a pendulum is analogous to planning on a 'staircase' branch within the configuration space. For further details, we refer the reader to \cite{campolo2023quasi}.

\begin{figure}[H]
\centering
\includegraphics[width=0.4\textwidth]{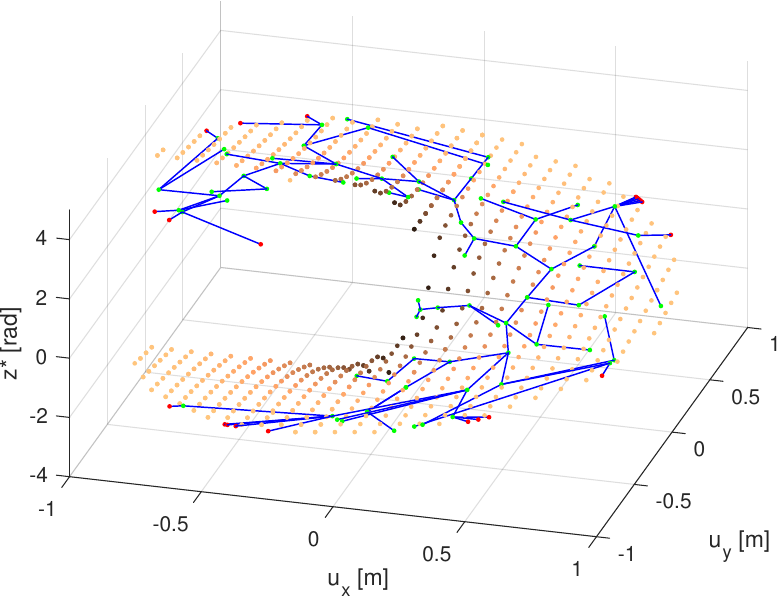}
\caption{HapticRRT navigates on one branch of $\mathcal{M}^{eq}$, where green nodes represents stable state, red denotes unstable states (haptic obstacle). }
\label{fig: tree on staircase}
\end{figure}

In Fig. \ref{fig: tree on staircase}, we set the maximum number of nodes to $N = 100$ for HapticRRT. The underlying manifold, as identified by \cite{campolo2023quasi}, is depicted in orange, serving as a backdrop for our analysis. The nodes of HapticRRT tree are represented by green points, while the edges connecting these nodes are shown as blue straight lines.
Notably, when exploration begins from the 'staircase' branch of the manifold, HapticRRT efficiently expands within this branch. Meanwhile, the red point marks where the ODE is terminated due to the presence of singularity, i.e., haptic obstacle (Eq. \ref{eq: Haptic Obstacle}).
This phenomenon commonly occurs when a node approaches the boundary of the branch or when the path leads to instability. As the node nears the boundary of the branch, it may transition into an unstable state, analogous to a scenario where a robot is holding a pendulum but suddenly releases it, leading to loss of control.

\vspace{-5mm}
\subsection{Visualization of Haptic Metric}
\begin{figure}[!ht]
\centering
\includegraphics[width=0.4\textwidth]{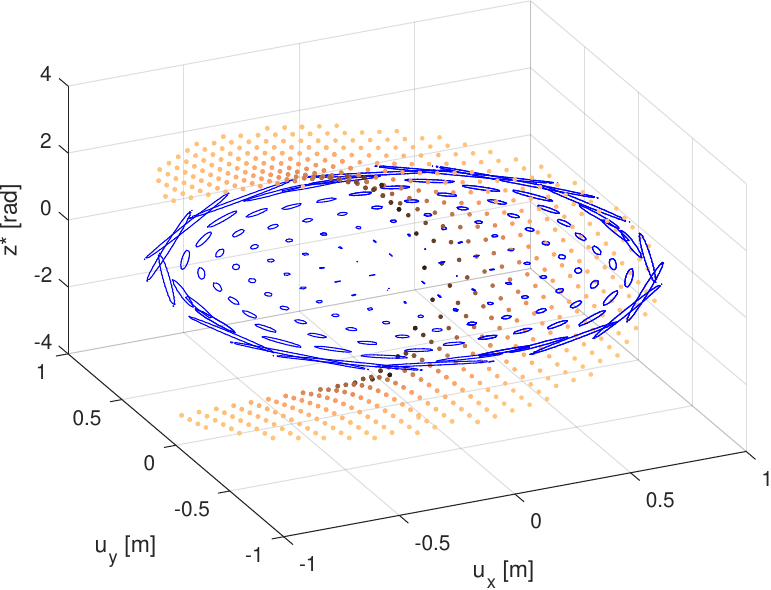}
\caption{Haptic metric in control space $\mathcal U$ for the example of pendulum, while blue ellipse represents haptic metric.}
\label{fig: metric}
\end{figure}

To better understand the concept of haptic metric, we visualize it as a blue ellipse, defined by the equation: $\V u^T \V G^2_m(\V z^*(\V u), \V u) \V u = 1$. This ellipse is plotted in the control space ($u_x, u_y$ in this case), as shown in Fig. \ref{fig: metric}.

The size of the ellipse reflects the eigenvalues of the haptic metric, while the orientation of the ellipse provides further insights:
\begin{itemize}
    \item The long axis of ellipse corresponds to smaller eigenvalues, indicating that manipulation in that direction requires less force. Thus, pushing the pendulum along the tangent direction at the tip requires less force.
    \item Conversely, the short axis represents the higher eigenvalue, indicating that squeezing the pendulum (applying force along to its length) requires more force.
    \item Near the outer boundary of the staircase, the ellipses are larger, suggesting that manipulating the pendulum is easier at its tip than at its origin.
\end{itemize}

\subsection{Experiment}

We validate our method on a real world setup, as shown in Fig. \ref{fig: inverted pendulum exp}. 
A robot with a circular finger continuously pushes a hinged pendulum to rotate it toward a target configuration. 
The key challenge of this task is to sustain contact while adapting both the pushing direction and force according to the pendulum's configuration and mass.

We compare the proposed \textit{HapticRRT} with \textit{AtlasRRT}, implemented using the OMPL library \cite{kingston2019exploring}, where the constraint equation for the inverted pendulum encodes its geometric constraints, and an external pushing force is manually specified.
Theoretically, AtlasRRT does not take mass into account, as its constraint model is purely kinematic. 
Therefore, when the weight of the pendulum changes, it cannot infer the required amount of force. 

As shown in Table \ref{table:IP}, when the pendulum is light (0.1~kg), both methods perform well.
However, when the pendulum is heavy (0.5~kg), AtlasRRT fails while our method succeeds. 
This highlights a key difference between motion planning and manipulation planning where our framework incorporates contact and gravity into the manipulation potential, allowing HapticRRT to reason about the need for sustained pushing and to adapt the force accordingly, therefore spend longer time.

\begin{table}[!h]
\renewcommand{\arraystretch}{1}
\caption{Success rate and planning time under different \textbf{pendulum masses}.\label{table_result}}
\centering
\footnotesize
\begin{tabular}{cccc}
\toprule
\textbf{Method} & \textbf{0.1 kg pendulum} & \textbf{0.5 kg pendulum} & \textbf{Time (s)} \\
\midrule
\textbf{AtlasRRT \cite{kingston2019exploring}} & 4/5 & 0/5 & 0.076 \\
\textbf{HapticRRT} & 5/5 & 5/5 & 0.305 \\
\bottomrule
\end{tabular}
\label{table:IP}
\end{table}

\section{Manipulation of Spring-Loaded Clip} \label{sec:clip}

The next manipulation task involves operating a spring-loaded clip and clipboard. 
This task requires sequential execution: the object can only be inserted after the clip has been successfully opened. 
Moreover, the required force to open the clip varies depending on the object's size and the stiffness of the clip.
We apply \textit{HapticRRT} to this task to reason about contact forces and strategy.

\begin{figure}[H]  
  \centering
  \begin{subfigure}{0.19\textwidth}
    \includegraphics[width=\linewidth]{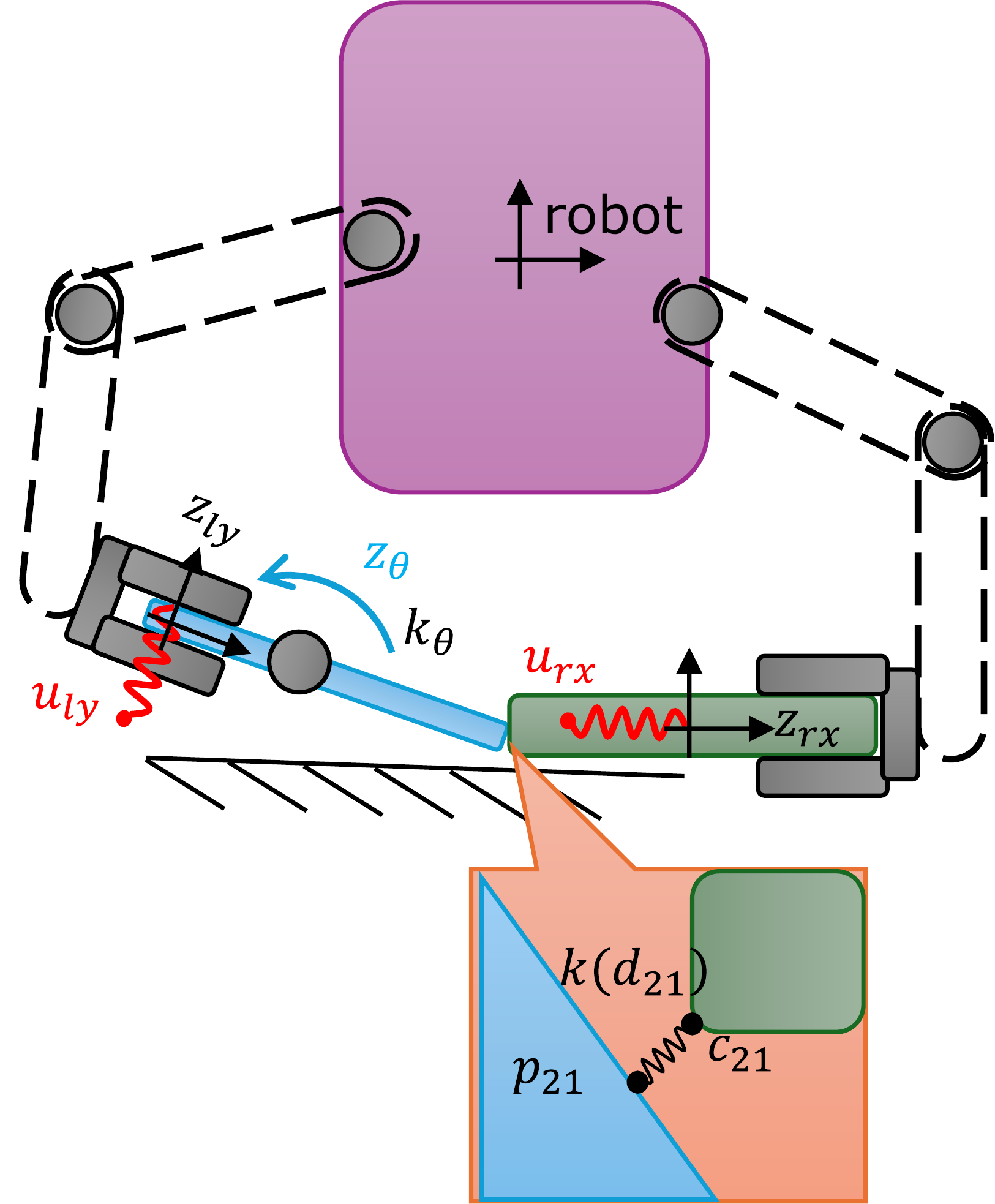}
    \caption[]
    {\small System modeling.}
    \label{fig:clipper_model}
  \end{subfigure}
  \hspace{2mm}
  \begin{subfigure}{0.27\textwidth}
    \includegraphics[width=\linewidth]{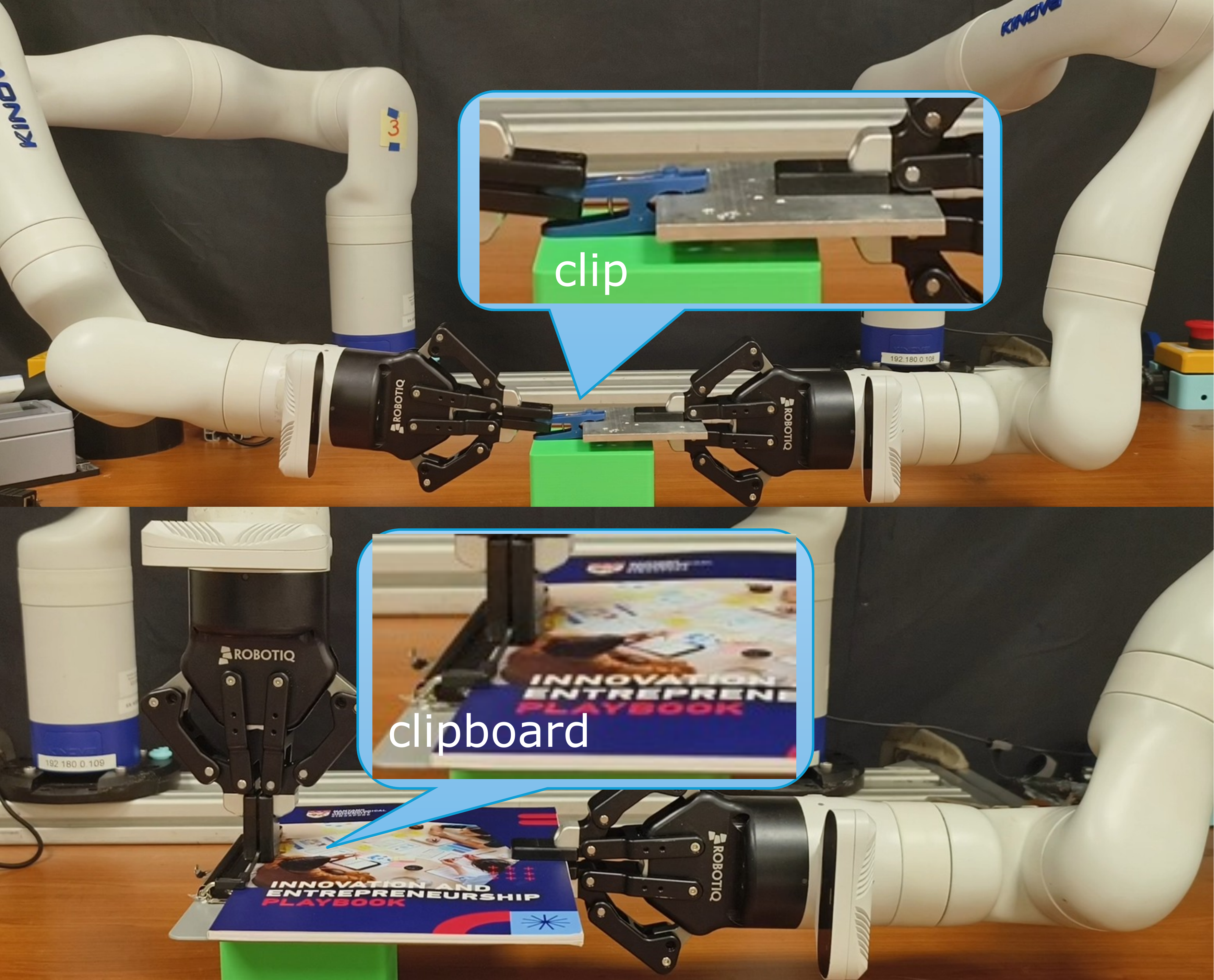}
    \caption[] 
    {\small Real world setup: spring-loaded clip and clipboard.}
    \label{fig:clipper_real}
  \end{subfigure}
  \caption{Manipulating a spring-loaded clip with varying clip type and object size.} 
  \label{fig:clip}
\end{figure}

\vspace{-7mm}
\subsection{Clip System Modeling}

We model the contact interaction between the object and the clip using a nonlinear stiffness function $k(d)$ (Eq.\ref{eq: stiffness}), along with proxies representing contact points. 
For details of the proxy modeling approach, we refer the reader to \cite{yang2025planning_book}. 
In brief, the proxy parameterizes the contact location on a SQ surface that is closest to the manipulated object, formulated as:
\begin{align}
    \underset{\gamma }{\arg \min } \quad & \norm{\V c(\V z) - \V p(\gamma)}, \quad \gamma \in [0, 2\pi] \label{eq:proxy}
\end{align}
where $\V p(\gamma)$ denotes the position of the proxy on the clip, and $\V c(\V z)$ denotes the relevant corner point of the object. We have multiple proxies to capture all the contact points.

We define the manipulation potential as illustrated in Fig. \ref{fig:clipper_model}. 
To simplify the representation, we define the control and state variables as $\V u = [u_{ly}, u_{rx}]$ and $\V z = [z_\theta, \V z_r]$, where $\V z_r = [z_{ly}, z_{rx}]$ denotes the positions of the dual arm end effectors.
In addition, the clip includes a rotational stiffness term $k_\theta$ that resists its opening.

The overall manipulation potential is defined as:
\begin{align}\label{eq: W_tot_cliper}
{W}(\V z^*, \V u) & =  W_\text{ctrl}+W_\text{clip} +W_\text{contact} \nonumber \\
&= \frac{1}{2}  (\V u - \V z_r)^T \V K_c (\V u - \V z_r) + \frac{1}{2} k_\theta (\V z_\theta - \V z_{\theta ,0})^2 \nonumber \\
&+ \sum_i \sum_j \frac{1}{2} k(d_{ij}) \|\V c_{ij} - \V p_{ij} \|^2
\end{align}
This potential consists of three components:
$W_\text{ctrl}$ represents the control energy applied by the robot,
$W_\text{clip}$ captures the rotational resistance of the clip,
and $W_\text{contact}$ models the contact energy between the object and the clip.

It is worth noting that the only difference between the classical clip and the clipboard task lies in the grasping position of the robot's left arm and pushing direction.
Therefore, the manipulation potential formulation remains the similar for both cases.

\subsection{Multiple Branches Discovery}

\begin{figure}[!h]  
    \centering

    \begin{subfigure}{0.241\textwidth}
        \includegraphics[width=\linewidth]{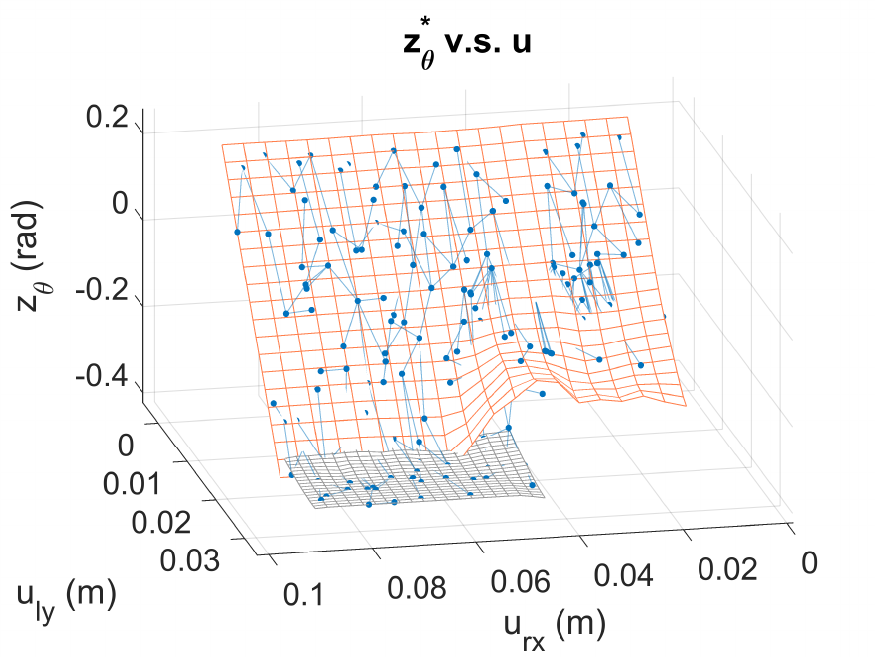}
        \caption[]
        {\small Multiple branches of $\mathcal M^{eq}$.}
        \label{SUBFIG:bkclipper_zvsu}
    \end{subfigure}
    \hfill
    \begin{subfigure}{0.241\textwidth}
        \includegraphics[width=\linewidth]{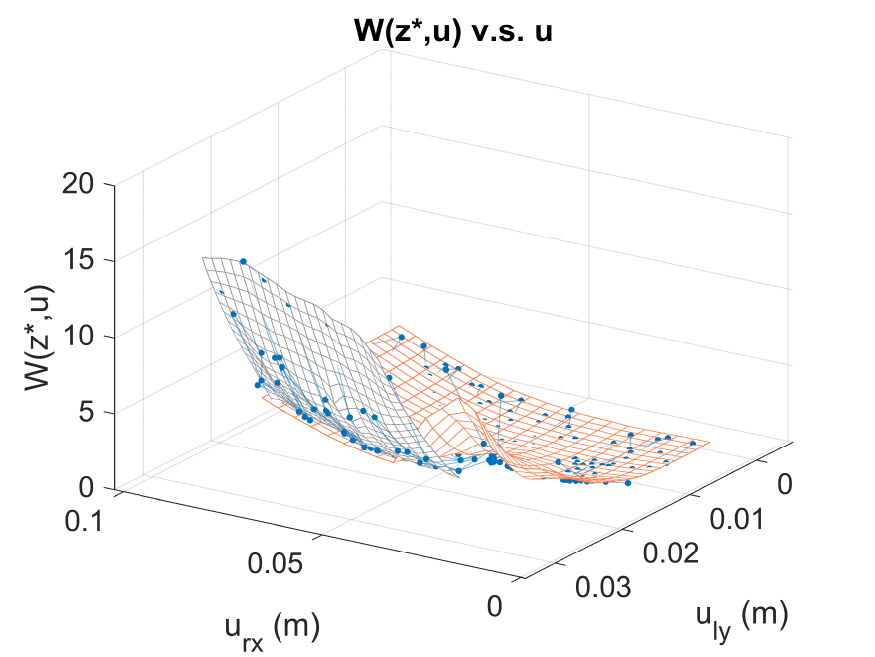}
        \caption[] 
        {\small $W(\V z^*, \V u)$ across branches.}
        \label{SUBFIG:bkclipper_Wvsu}
    \end{subfigure}
    \vskip\baselineskip  
    \begin{subfigure}{0.241\textwidth}
        \includegraphics[width=\linewidth]{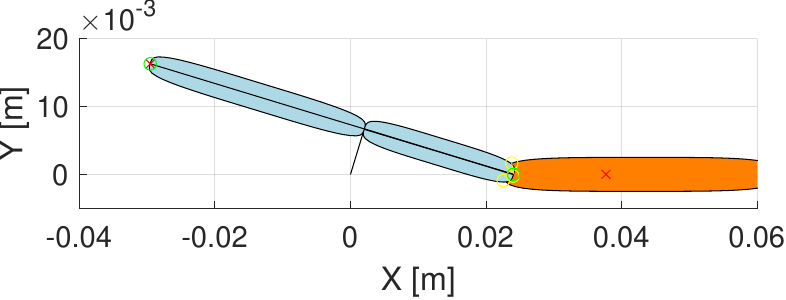}
        \caption[] 
        {\small Grey mesh: stuck.}
        \label{SUBFIG:bkclipper_3}
    \end{subfigure}
    \hfill
    \begin{subfigure}{0.241\textwidth}
        \includegraphics[width=\linewidth]{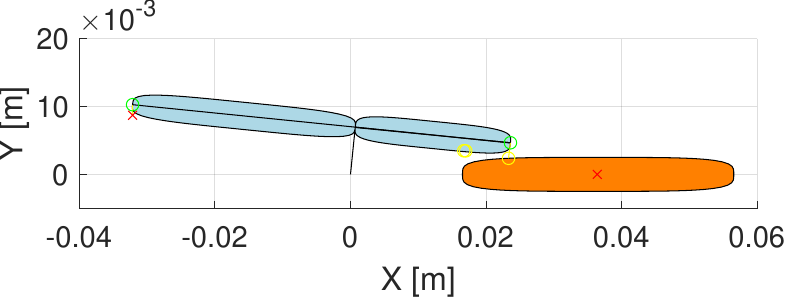}
        \caption[] 
        {\small Orange mesh: success.}
        \label{SUBFIG:bkclipper_4}
    \end{subfigure}

    \caption{These mesh plots show the discovered branches. Grey mesh corresponds to incorrect operation sequences where the object becomes stuck in front of the clip (Fig. \ref{SUBFIG:bkclipper_3}). Orange mesh represents successful insertions following the correct sequence (Fig. \ref{SUBFIG:bkclipper_4}).}
    \label{fig:meshclipper}
\end{figure}

Branches emerge when two objects push against each other, allowing the manipulated object to slide to either side of the other object \cite{poston2014catastrophe}. 
We apply \textit{HapticRRT} to explore $\mathcal{M}^{eq}$ in clip scenario, with results shown in Fig. \ref{fig:meshclipper}.

In this example, two distinct branches indicate:
\begin{itemize}
    \item \textbf{Grey branch:} This branch corresponds to the robot pushing the object before opening the clip. In Fig. \ref{SUBFIG:bkclipper_zvsu}, the grey mesh is located where $u_{ly}$ is large (the left arm does not push the clip), and $z_{\theta}$ remains around $-0.4$ (the clip remains closed). As a result, $u_{rx}$ does not exceed 0.4, indicating that the object cannot enter the clip. In Fig. \ref{SUBFIG:bkclipper_Wvsu}, the grey mesh exhibits higher potential values, consistent with physical resistance.
    \item \textbf{Orange branch:} This branch corresponds to first opening the clip and then inserting the object. In Fig. \ref{SUBFIG:bkclipper_zvsu}, the orange mesh appears when $u_{ly}$ is close to zero (the robot opens the clip), and $z_{\theta}$ increases accordingly. Consequently, $u_{rx}$ approaches to zero, meaning the object successfully enters the clip. This successful behavior is also reflected in Fig. \ref{SUBFIG:bkclipper_Wvsu}, where the orange mesh has lower potential values.
\end{itemize}

\vspace{-3mm}
\subsection{Comparison with Prior Method} \label{sec:sompare_clip}

Classical motion planners such as AtlasRRT \cite{kingston2019exploring} are inadequate for tasks involving contact-rich manipulation, as they do not model the state of passive objects or the required contact forces.

To address such limitations, learning-based approaches (e.g., reinforcement learning (RL) and evolutionary strategies (ES)) are often employed. In our prior work \cite{yang2025planning_book}, we proposed a policy optimization framework that combines Dynamic Movement Primitives (DMPs) with black-box optimization (BBO). This method can be viewed as a form of policy search, conceptually related to REINFORCE and ES \cite{stulp2013robot}.

In this section, we compare the proposed HapticRRT with our previous BBO method. As shown in Table \ref{table_DMPBBO_clip}, HapticRRT significantly reduces the required computation time. This result highlights the efficiency of tree-based planner over iterative optimization. Meanwhile, hapticRRT achieves optimality within the tree structure, though not necessarily global optimality. Hence hapticRRT has a larger haptic distance $\phi$.

\begin{table}[H]
\renewcommand{\arraystretch}{1}
\caption{Comparison of HapticRRT with a prior optimization-based approach.\label{table_DMPBBO_clip}}
\centering
\footnotesize  
\begin{tabular}{ccc}
\toprule
\textbf{Method} & Computing time & Haptic distance \\
\midrule
\textbf{DMP-BBO \cite{yang2025planning_book}} & 16.14 s & 23.92 \\
\midrule
\textbf{HapticRRT} & 2.74 s & 27.23 \\
\bottomrule
\end{tabular}
\label{table:IP}
\end{table}

\vspace{-5mm}

\subsection{Experiment: Real World Validation}

We validate our method in four real world cases: three involving spring-loaded clothespins with different object sizes, and one involving a clipboard. 
In each case, the robot uses a two-finger gripper to grasp one side of the clip, while the other side is placed against a table to enable non-prehensile manipulation, thus avoiding reliance on a dexterous hand. Another robot grasp the object to insert as Fig.\ref{fig:clipper_real}.
\begin{table}[H]
\renewcommand{\arraystretch}{1}
\caption{Successful rate under different condition.\label{table_clip}}
\centering
\resizebox{\linewidth}{!}{
\begin{tabular}{ccccc}
\toprule
\textbf{Clip Type} & clothespin & clothespin & clothespin & clipboard \\
\midrule
\textbf{Object size} & 5 mm & 3 mm & 1 mm & 5 mm \\
\midrule
\textbf{Success Rate} & 5/5 & 5/5 & 5/5 & 4/5 \\
\bottomrule
\end{tabular}
}
\end{table}
The results are summarized in Table \ref{table_clip}. 
For each setting, we repeat the execution of the output policy five times in the real world. All clothespin cases achieve success, while the clipboard case has a single failure (4/5 success). 
This demonstrates that HapticRRT produces robust and repeatable behavior for contact-rich manipulation tasks.


\begin{figure}[H]
\centering
\includegraphics[width=0.4\textwidth]{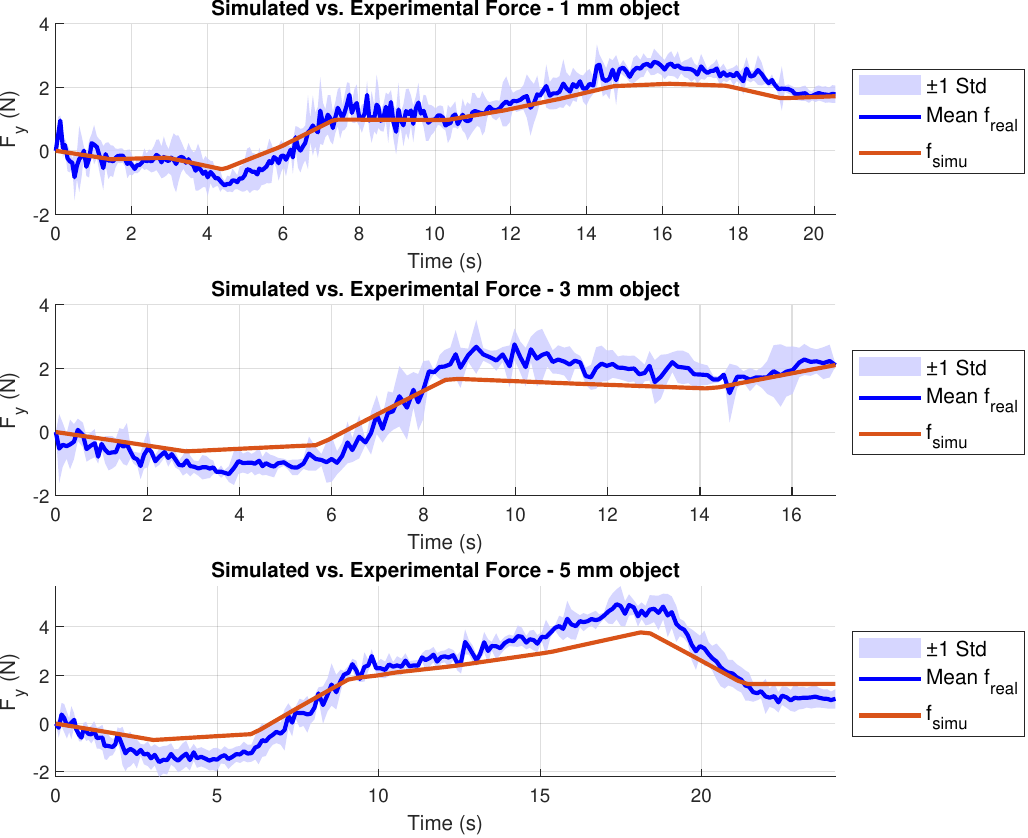}
\caption{Simulation v.s. experiment: External force on left arm during clip manipulation.}
\label{fig:force_clip}
\end{figure}

We also compare the predicted contact force with real world data for the three clothespin cases. 
The force data is collected from the Kinova joint torque sensors, and post-processed to estimate external contact force. 
As shown in Fig. \ref{fig:force_clip}, the blue lines indicate the mean and variance over five experimental trials, while the red lines represent the predicted force ($-\partial_u W$) from our framework.

The force profiles closely match. 
At the beginning, the left arm applies near-zero force, as opening the clip is unnecessary when the object is still far away, which conserves energy.
As the object approaches the clip, HapticRRT increases the left arm's pushing force to open the clip appropriately.
Among all cases, the 5~mm object requires the highest force, as the clip must open the widest to allow insertion.

\section{Crowded bookshelf insertion}\label{sec:book}

Building upon our previous work \cite{yang2025planning_book}, we apply HapticRRT to a contact-rich task: inserting a book into a crowded shelf where the available space is insufficient for direct insertion.
To complete the task, the robot must first push neighboring books aside before inserting the new one.
\begin{figure}[H]  
  \begin{subfigure}{0.245\textwidth}
    \includegraphics[width=\linewidth]{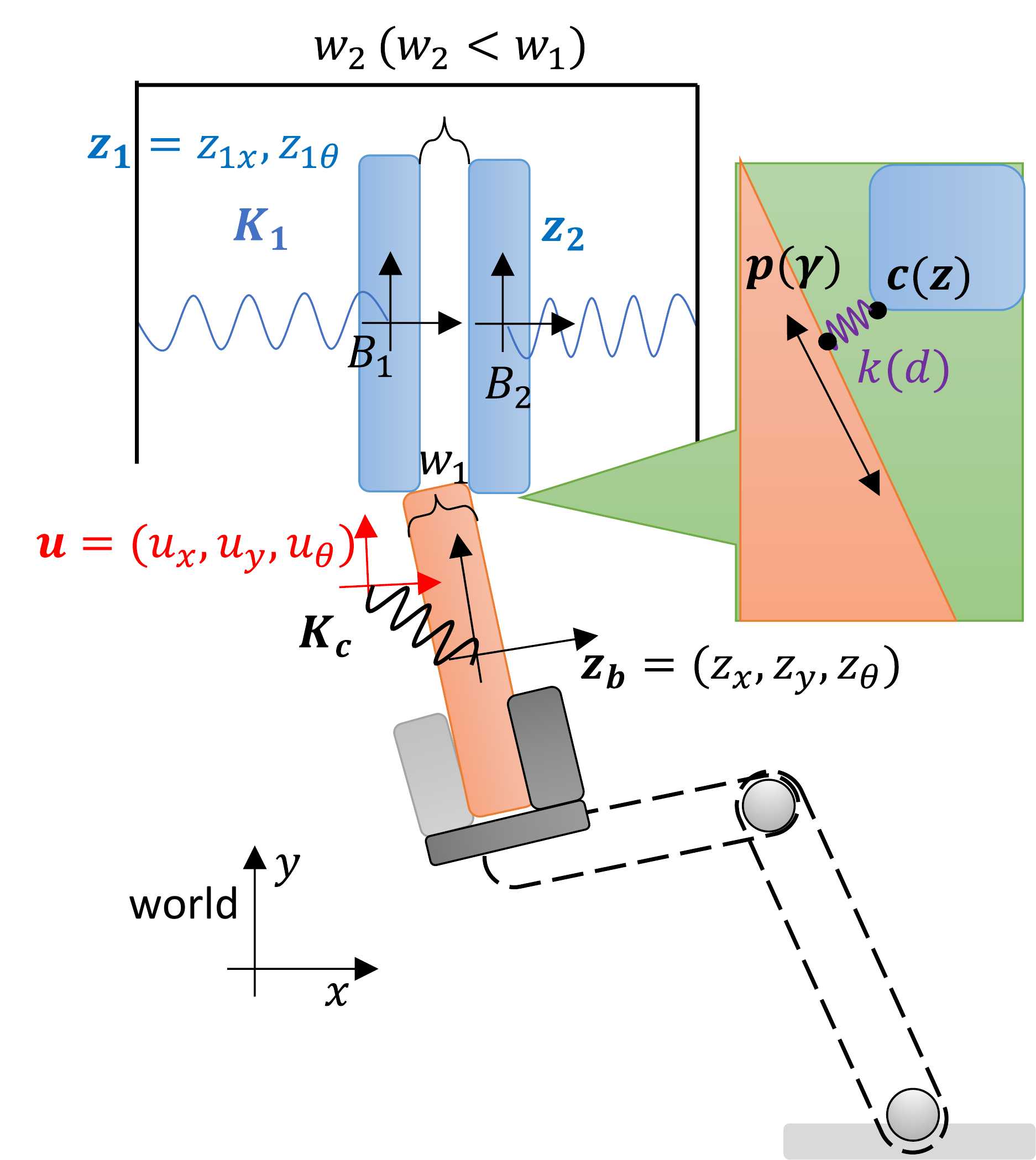}
    \caption[]
    {\small Modeling of book insertion.}
    \label{SUBFIG:bookmodel}
  \end{subfigure}%
  \hfill  
  \begin{subfigure}{0.24\textwidth}
    \includegraphics[width=\linewidth]{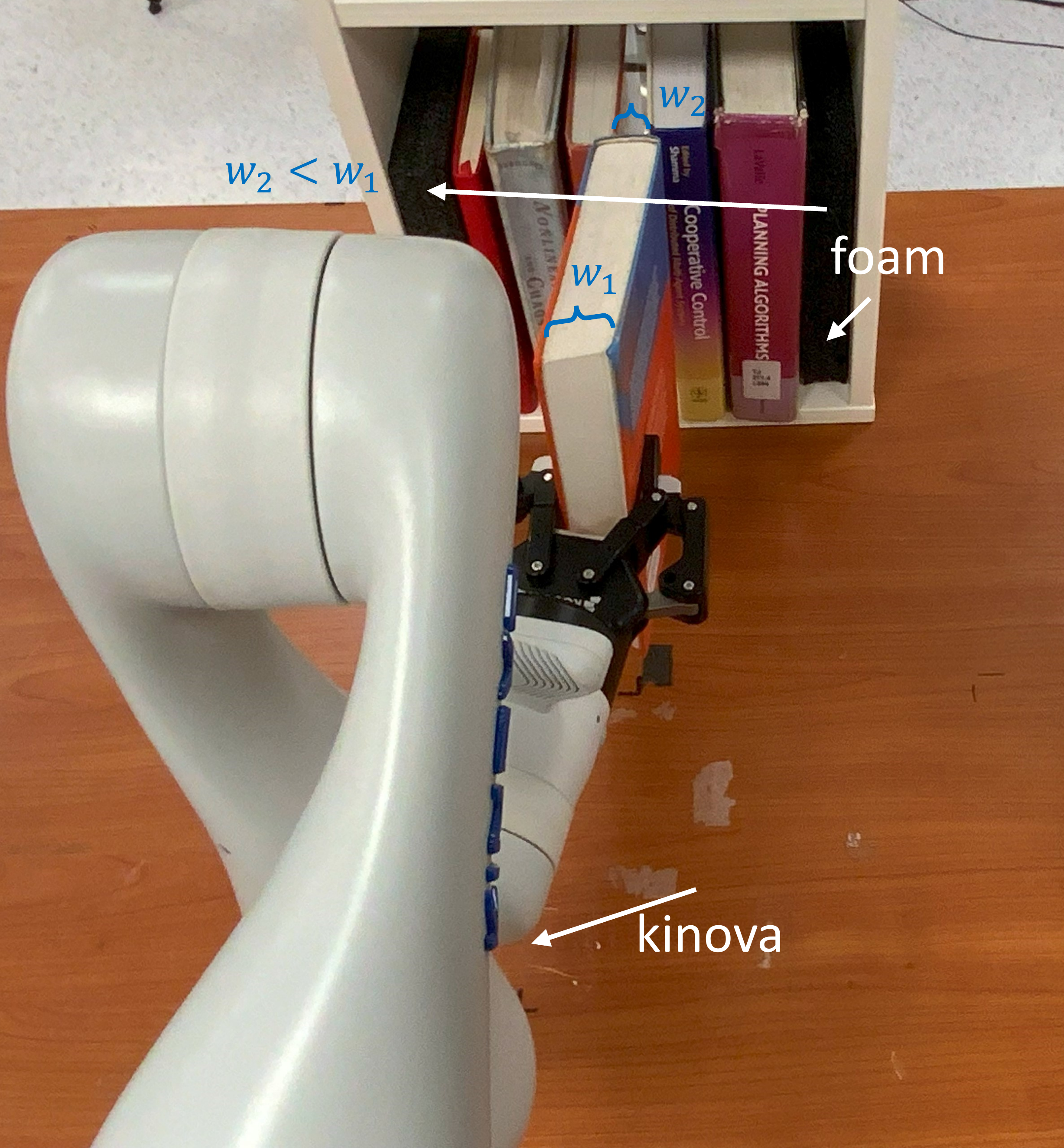}
    \caption[] 
    {\small Experimental setup.}
    \label{SUBFIG:real}
  \end{subfigure}%
\caption{Modeling and experimental setup of the bookshelf insertion. 
The book $\V z_b$ is inserted into a narrow space ($w_2 < w_1$) under contact and resistance from neighboring books.} 
\end{figure}

\subsection{Crowded Shelf Modeling}

We reuse the modeling framework from \cite{yang2025planning_book}, as illustrated in Fig. \ref{SUBFIG:bookmodel}. 
The robot manipulates the book in a planar space, with control input $\V u = [u_x, u_y, u_\theta]^T \in SE(2)$ and book state $\V z_b = [z_x, z_y, z_\theta]^T$.
Two neighboring books, $\V z_1$ and $\V z_2$, are modeled as passive bodies connected to virtual springs with stiffness matrices $\V K_1$ and $\V K_2$, and rest positions $\V z_{i,0}$.
The gripper uses impedance control with stiffness matrix $\V K_c$.
As in prior sections, contact interactions are modeled using proxy $\gamma$, and the overall manipulation potential is defined as:
\begin{align}\label{eq: W_tot_book}
{W}(\V z^*, \V u) & =  W_\text{ctrl}+W_\text{resist} +W_\text{contact} \nonumber \\
&= \frac{1}{2}  (\V u - \V z_b)^T \V K_c (\V u - \V z_b) \nonumber \\
&+\sum_{i=1,2}  \frac{1}{2} (\V z_i - \V z_{i,0})^T \V K_i (\V z_i - \V z_{i,0})  \nonumber \\
&+ \sum_i \sum_j \frac{1}{2} k(d_{ij}) \|\V c_{ij} - \V p_{ij} \|^2
\end{align}
This potential consists of three terms:
$W_\text{ctrl}$ is the control energy from the impedance control,
$W_\text{resist}$ captures the passive resistance of the neighboring books,
and $W_\text{contact}$ models contact interactions among books.

\subsection{Exploring Equilibrium Manifold}

\begin{figure}[H]  
    \centering
    \begin{subfigure}{0.243\textwidth}
        \centering
        \includegraphics[width=\linewidth]{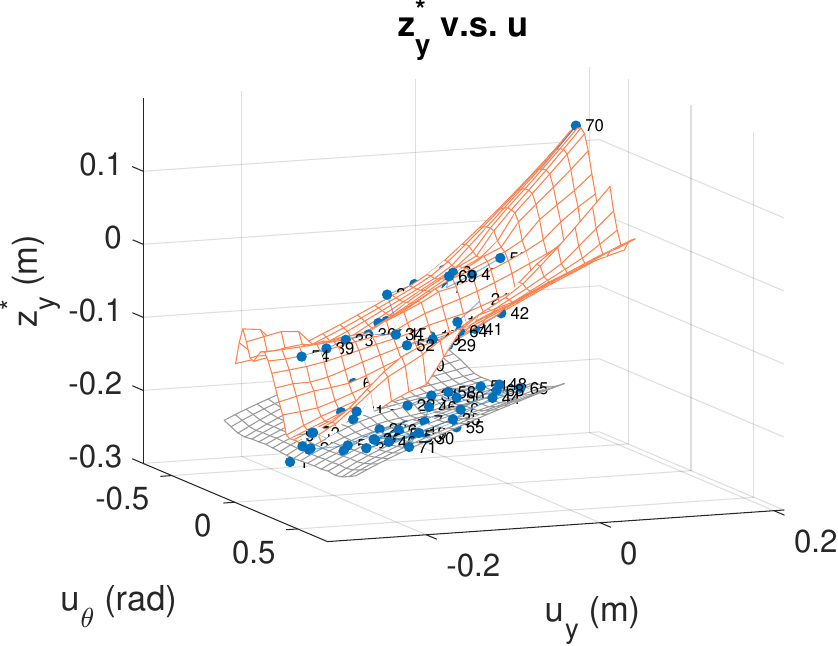}
        \caption[]
        {\small HapticRRT discovers $\mathcal M^{eq}$.}
        \label{SUBFIG:zy_u}
    \end{subfigure}%
    \hfill 
    \begin{subfigure}{0.243\textwidth}
        \centering
        \includegraphics[width=\linewidth]{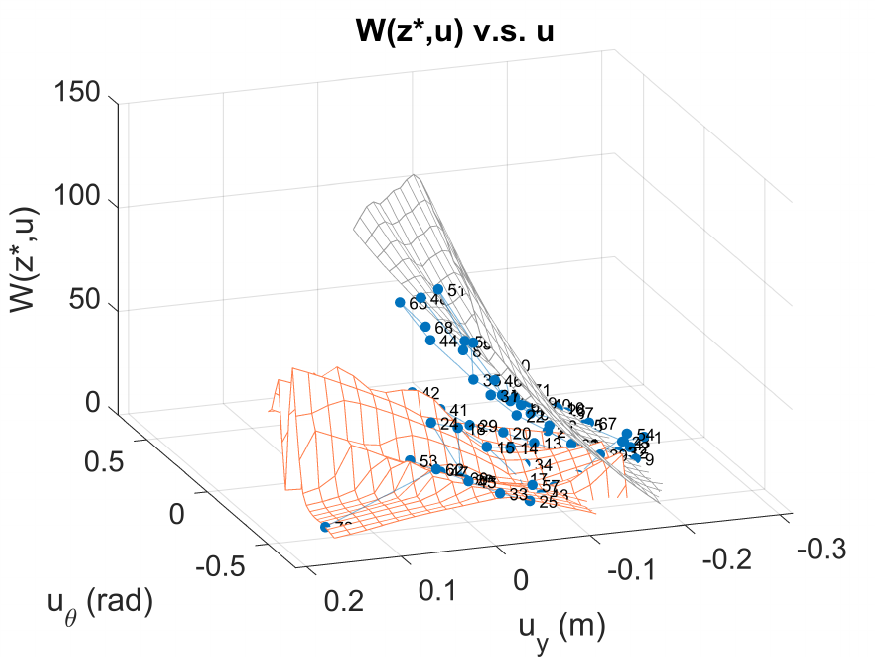}
        \caption[] 
        {\small $W(\V z^*, \V u)$ across manifold.}
        \label{SUBFIG:W_u}
    \end{subfigure}%
    \caption{HapticRRT explores $\mathcal M^{eq}$ and reveals distinct insertion strategies.}
    \label{fig:meshtree}
\end{figure}

We apply HapticRRT to explore $\mathcal M^{eq}$ in this bookshelf insertion task. 
Fig. \ref{fig:meshtree} illustrates the resulting mesh representations and the exploration tree. 
Specifically, we visualize $z_y$ against control inputs $u_\theta$ and $u_y$ in Fig. \ref{SUBFIG:zy_u}, and the corresponding manipulation potential $W(\V z^*, \V u)$ in Fig. \ref{SUBFIG:W_u}.

We select $z_y$ as the vertical axis in Fig. \ref{SUBFIG:zy_u}, since $z_y = 0$ corresponds to a fully inserted book.
In the grey mesh, $z_y$ remains flat as $u_y$ increases, indicating that the book is getting stuck in front of the neighboring books due to insufficient space.
In contrast, the orange mesh represents a different strategy discovered by HapticRRT, where the robot first shifts the neighboring books to create space before inserting the target book. As a result, $z_y$ increases significantly, indicating successful insertion.
A similar trend is observed in Fig. \ref{SUBFIG:W_u}.
When the robot pushes forward without addressing the environmental constraints, the manipulation potential $W(\V z^*, \V u)$ increases continuously.
In contrast, once HapticRRT discovers wedging-in policy, the potential decreases, suggesting that the task has been successfully executed.

\vspace{-2mm}
\subsection{Comparison on Book Insertion}

As discussed in Section \ref{sec:sompare_clip}, we also compare HapticRRT with the DMP-BBO approach on the crowded book insertion task.

As shown in Table \ref{table_DMPBBO}, HapticRRT again achieves significantly lower computation time, indicating superior planning efficiency. However, since this task is more complex than the previous one, the BBO method benefits from a longer optimization time, resulting in a lower haptic distance $\phi$ due to its ability to explore global optimal solutions.
\begin{table}[!h]
\renewcommand{\arraystretch}{1}
\caption{Compare on book insertion.\label{table_DMPBBO}}
\centering
\footnotesize  
\begin{tabular}{ccc}
\toprule
\textbf{Method} & Computing time & Haptic distance \\
\midrule
\textbf{DMP-BBO \cite{yang2025planning_book}} & 791.22 s & 22.06 \\
\midrule
\textbf{HapticRRT} & 48.71 s & 35.78 \\
\bottomrule
\end{tabular}
\label{table:IP}
\end{table}


\subsection{Experiment: Real World Validation}

The experimental setup is shown in Fig.~\ref{SUBFIG:real}. 
Foam sheets are attached to both sides of the bookshelf to simulate stiffness, and several books are placed to leave a narrow slot of width $w_2$. 
A Kinova Gen3 robot grasps a book of width $w_1 > w_2$, making direct insertion infeasible.
To evaluate robustness, we vary both the book width and its initial position across trials.

\begin{table}[!h]
\renewcommand{\arraystretch}{1}
\caption{Successful rate under different condition.\label{table_result}}
\centering
\resizebox{\linewidth}{7mm}{
\begin{tabular}{ccccccc}
\toprule
\textbf{Experiment Type} & initial pose & initial pose & initial pose & initial pose & initial pose & book size \\
\midrule
\textbf{Variation} & x=-0.05 m & x=-0.025 m & x= 0 m & x=0.025 m & x=0.05 m & increased\\
\midrule
\textbf{Success Rate} & 5/5 & 5/5 & 4/5 & 5/5 & 4/5 & 4/5\\
\bottomrule
\end{tabular}}
\end{table}

In most cases, the control trajectory from HapticRRT successfully executed the task. However, some failures occurred due to the jagged and non-optimal nature of the trajectory, leading to excessive force application. In some cases, the book was pushed too hard, causing deformation and slippage, which resulted in failure.

\begin{figure}[!h]  
\centering
  \begin{subfigure}{0.41\textwidth}
    \includegraphics[width=\linewidth]{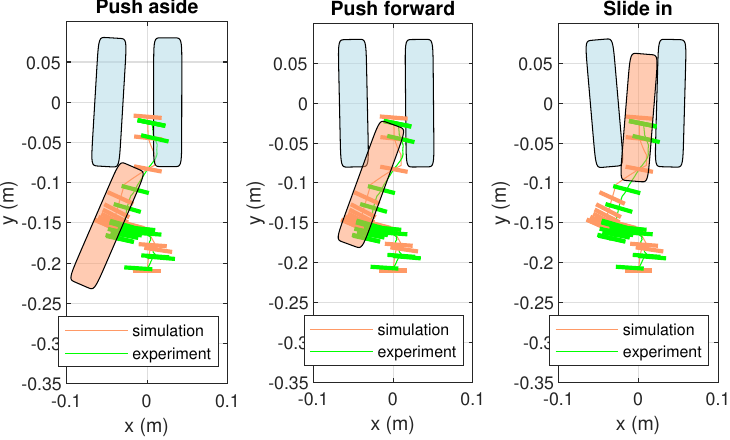}
    \caption[]
    {\small Simulation v.s. experiment: trajectory of the book $\V z(t)$ during the insertion process.}
    \label{SUBFIG:phase}
  \end{subfigure}%
  \vfill  
  \begin{subfigure}{0.42\textwidth}
    \includegraphics[width=\linewidth]{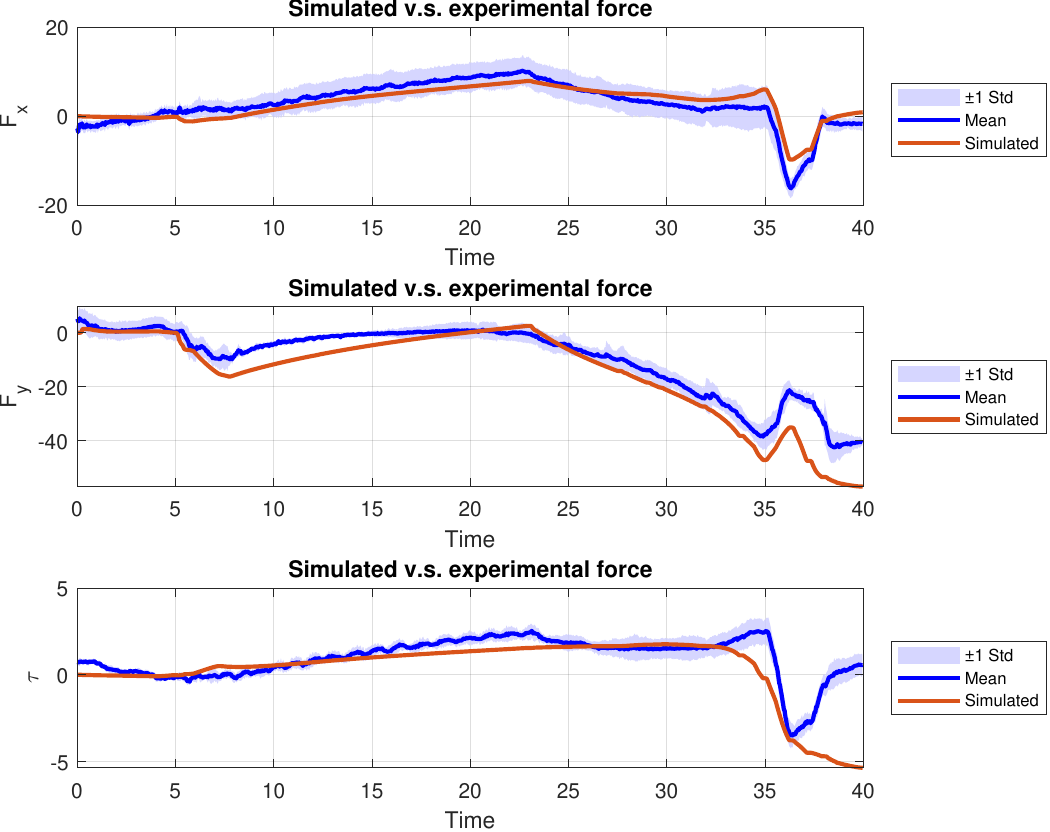}
    \caption[] 
    {\small  Simulation v.s. experiment: External wrench during the insertion process.}
    \label{SUBFIG:force}
  \end{subfigure}%
\caption{Real-world implementation of HapticRRT: trajectory and force comparison.} 
\label{fig: exp_vs_simu}
\end{figure}
One typical insertion policy and its real-world implementation are shown in Fig. \ref{fig: exp_vs_simu}. The book trajectory $\V z(t)$ in both the simulation and experiment are plotted in orange and green, respectively, with the short lines indicating the book's orientation.
In the experiment, the contact force is computed from the external torque reading from robot joint sensor, and adjusted to account for the weight of the manipulated book. The simulated contact force (red curve) is defined as $-\partial_{\V u} W$. 
Similar to previous analysis, the blue lines indicate the mean and variance over five experimental trials. HapticRRT automatically discovers an interpretable three-phase insertion strategy after initial contact:
\begin{itemize}
    \item \textbf{Push aside:} The robot applies strong lateral force ($F_x$, $\tau$) to shift the neighboring book and create space. 
    \item \textbf{Push forward:} Once sufficient space is available, the robot begins insertion. The forward force $F_y$ increases, reflecting resistance along the insertion axis.
    \item \textbf{Slide in:} As the book enters the shelf, resistance decreases and $F_x$ converges. However, $F_y$ and $\tau$ remain non-zero, since HapticRRT does not optimize for minimal force, and may apply excess effort after successful insertion.
\end{itemize}
The force trends and magnitudes in both simulation and real world trials show strong consistency, validating the effectiveness of HapticRRT in contact-rich manipulation.

\section{Conclusion}

In this work, we proposed HapticRRT, a haptic sampling-based motion planning algorithm within a novel manipulation framework. By integrating classical motion planning into contact-rich manipulation, our method successfully discovers multiple branches of the equilibrium manifold and finds feasible solutions for contact-rich tasks.
We validated our approach in various tasks: pendulum manipulation, crowded bookshelf insertion and clip manipulation. Through these experiments, we visualized the physical meaning of haptic metrics and haptic obstacles, demonstrating the interpretability of our framework. 
Compared to classical motion planners, and our prior approach, HapticRRT demonstrates higher planning efficiency across diverse settings. 
The results demonstrate the robustness of HapticRRT, achieving a high success rate across varying conditions. Additionally, real-world experiments confirmed that the observed policy aligns well with simulation, proving the reliability of our framework.
More importantly, this work bridges the gap between collision-free motion planning and manipulation planning, showcasing its broad potential for real-world applications.
Future directions include improving sampling efficiency and developing an online adaptation mechanism using force feedback for real-time adjustments.

\section*{Acknowledgment}
		
This research is supported by the National Research Foundation, Singapore, under the NRF Medium Sized Centre scheme (CARTIN).
        
We would like to express our sincere gratitude to Donghan Yu for his insightful discussions and technical suggestions during the early stage of this work.

\bibliographystyle{Bibliography/IEEEtranTIE}
\bibliography{ref}\ 

\begin{thebibliography}{10}
\providecommand{\url}[1]{#1}
\csname url@samestyle\endcsname
\providecommand{\newblock}{\relax}
\providecommand{\bibinfo}[2]{#2}
\providecommand{\BIBentrySTDinterwordspacing}{\spaceskip=0pt\relax}
\providecommand{\BIBentryALTinterwordstretchfactor}{4}
\providecommand{\BIBentryALTinterwordspacing}{\spaceskip=\fontdimen2\font plus
\BIBentryALTinterwordstretchfactor\fontdimen3\font minus \fontdimen4\font\relax}
\providecommand{\BIBforeignlanguage}[2]{{%
\expandafter\ifx\csname l@#1\endcsname\relax
\typeout{** WARNING: IEEEtran.bst: No hyphenation pattern has been}%
\typeout{** loaded for the language `#1'. Using the pattern for}%
\typeout{** the default language instead.}%
\else
\language=\csname l@#1\endcsname
\fi
#2}}
\providecommand{\BIBdecl}{\relax}
\BIBdecl

\bibitem{suomalainen2022survey}
M.~Suomalainen, Y.~Karayiannidis, and V.~Kyrki, ``A survey of robot manipulation in contact,'' \emph{Robotics and Autonomous Systems}, vol. 156, p. 104224, 2022.

\bibitem{lavalle1998rapidly}
S.~LaValle, ``Rapidly-exploring random trees: A new tool for path planning,'' \emph{Research Report 9811}, 1998.

\bibitem{jimenez2024visualizing}
J.~O. Jimenez and W.~Suleiman, ``Visualizing high-dimensional configuration spaces: A comprehensive analytical approach,'' \emph{IEEE Robotics and Automation Letters}, 2024.

\bibitem{kingston2018sampling}
Z.~Kingston, M.~Moll, and L.~E. Kavraki, ``Sampling-based methods for motion planning with constraints,'' \emph{Annual review of control, robotics, and autonomous systems}, vol.~1, pp. 159--185, 2018.

\bibitem{jaillet2012path}
L.~Jaillet and J.~M. Porta, ``Path planning under kinematic constraints by rapidly exploring manifolds,'' \emph{IEEE Transactions on Robotics}, vol.~29, no.~1, pp. 105--117, 2012.

\bibitem{kingston2019exploring}
Z.~Kingston, M.~Moll, and L.~E. Kavraki, ``Exploring implicit spaces for constrained sampling-based planning,'' \emph{The International Journal of Robotics Research}, vol.~38, no. 10-11, pp. 1151--1178, 2019.

\bibitem{morgan2022complex}
A.~S. Morgan, K.~Hang, B.~Wen, K.~Bekris, and A.~M. Dollar, ``Complex in-hand manipulation via compliance-enabled finger gaiting and multi-modal planning,'' \emph{IEEE Robotics and Automation Letters}, vol.~7, no.~2, pp. 4821--4828, 2022.

\bibitem{zhou2023spatiotemporal}
Y.~Zhou, G.~Sun, Y.~Miao, Y.~Zhang, X.~Chen, and H.~Wang, ``Spatiotemporal optimal trajectory planning for safe planar manipulation of a moving object,'' \emph{IEEE Transactions on Industrial Electronics}, vol.~71, no.~7, pp. 7466--7476, 2023.

\bibitem{whitney1982quasi}
D.~E. Whitney \emph{et~al.}, ``Quasi-static assembly of compliantly supported rigid parts,'' \emph{Journal of Dynamic Systems, Measurement, and Control}, vol. 104, no.~1, pp. 65--77, 1982.

\bibitem{ozawa2017grasp}
R.~Ozawa and K.~Tahara, ``Grasp and dexterous manipulation of multi-fingered robotic hands: a review from a control view point,'' \emph{Advanced Robotics}, vol.~31, no. 19-20, pp. 1030--1050, 2017.

\bibitem{yang2023planning}
L.~Yang, M.~Z. Ariffin, B.~Lou, C.~Lv, and D.~Campolo, ``A planning framework for robotic insertion tasks via hydroelastic contact model,'' \emph{Machines}, vol.~11, no.~7, p. 741, 2023.

\bibitem{yang2025planning_book}
L.~Yang, S.~H. Turlapati, C.~Lv, and D.~Campolo, ``Planning for quasi-static manipulation tasks via an intrinsic haptic metric: A book insertion case study,'' \emph{IEEE Robotics and Automation Letters}, 2025.

\bibitem{CAMPOLO2025116003}
D.~Campolo and F.~Cardin, ``A geometric framework for quasi-static manipulation of a network of elastically connected rigid bodies,'' \emph{Applied Mathematical Modelling}, vol. 143, p. 116003, 2025.

\bibitem{campolo2023quasi}
D.~Campolo and F.~Cardin, ``Quasi-static mechanical manipulation as an optimal process,'' in \emph{2023 62nd IEEE Conference on Decision and Control (CDC)}, pp. 4753--4758.\hskip 1em plus 0.5em minus 0.4em\relax IEEE, 2023.

\bibitem{salem2020robotic}
A.~Salem and Y.~Karayiannidis, ``Robotic assembly of rounded parts with and without threads,'' \emph{IEEE Robotics and Automation Letters}, vol.~5, no.~2, pp. 2467--2474, 2020.

\bibitem{wang2024cooperative}
D.~Wang, C.~Qiu, J.~Lian, W.~Wan, Q.~Pan, and Y.~Dong, ``Cooperative control for dual-arm robots based on improved dynamic movement primitives,'' \emph{IEEE Transactions on Industrial Electronics}, 2024.

\bibitem{chen2024robust}
N.~Chen, L.~Wan, and Y.-J. Pan, ``Robust and adaptive dexterous manipulation with vision-based learning from multiple demonstrations,'' \emph{IEEE Transactions on Industrial Electronics}, 2024.

\bibitem{elguea2023review}
{\'I}.~Elguea-Aguinaco, A.~Serrano-Mu{\~n}oz, D.~Chrysostomou, I.~Inziarte-Hidalgo, S.~B{\o}gh, and N.~Arana-Arexolaleiba, ``A review on reinforcement learning for contact-rich robotic manipulation tasks,'' \emph{Robotics and Computer-Integrated Manufacturing}, vol.~81, p. 102517, 2023.

\bibitem{bing2022solving}
Z.~Bing, H.~Zhou, R.~Li, X.~Su, F.~O. Morin, K.~Huang, and A.~Knoll, ``Solving robotic manipulation with sparse reward reinforcement learning via graph-based diversity and proximity,'' \emph{IEEE Transactions on Industrial Electronics}, vol.~70, no.~3, pp. 2759--2769, 2022.

\bibitem{irfan2024control}
S.~Irfan, L.~Zhao, S.~Ullah, A.~Mehmood, and M.~Fasih Uddin~Butt, ``Control strategies for inverted pendulum: A comparative analysis of linear, nonlinear, and artificial intelligence approaches,'' \emph{Plos one}, vol.~19, no.~3, p. e0298093, 2024.

\bibitem{shaikh2023door}
J.~Shaikh-Mohammed, Y.~Alharbi, and A.~Alqahtani, ``Door-opening technologies: Search for affordable assistive technology,'' \emph{Technologies}, vol.~11, no.~6, p. 177, 2023.

\bibitem{kim2021integrated}
U.~Kim, D.~Jung, H.~Jeong, J.~Park, H.-M. Jung, J.~Cheong, H.~R. Choi, H.~Do, and C.~Park, ``Integrated linkage-driven dexterous anthropomorphic robotic hand,'' \emph{Nature communications}, vol.~12, no.~1, p. 7177, 2021.

\bibitem{nakajima2011study}
T.~Nakajima, T.~Yoshimi, M.~Mizukawa, and Y.~Ando, ``A study of book arrangement task by robot arm-book insert operation to bookshelf,'' in \emph{2011 IEEE/SICE International Symposium on System Integration (SII)}, pp. 738--743.\hskip 1em plus 0.5em minus 0.4em\relax IEEE, 2011.

\bibitem{sygo2023multi}
B.~Sygo, S.-C. Liu, F.~Wieczorek, M.~Koshil, M.~G{\"o}rner, N.~Hendrich, and J.~Zhang, ``Multi-stage book perception and bimanual manipulation for rearranging book shelves,'' in \emph{International Conference on Intelligent Autonomous Systems}, pp. 495--507.\hskip 1em plus 0.5em minus 0.4em\relax Springer, 2023.

\bibitem{spivak2018calculus}
M.~Spivak, \emph{Calculus on manifolds: a modern approach to classical theorems of advanced calculus}.\hskip 1em plus 0.5em minus 0.4em\relax CRC press, 2018.

\bibitem{yang2025energy}
L.~Yang, H.-T. Nguyen, C.~Lv, D.~Campolo, and F.~Cardin, ``An energy-based numerical continuation approach for quasi-static mechanical manipulation,'' \emph{Data-Centric Engineering}, vol.~6, p. e18, 2025.

\bibitem{jaklic2000segmentation}
A.~Jaklic, A.~Leonardis, and F.~Solina, \emph{Segmentation and recovery of superquadrics}, vol.~20.\hskip 1em plus 0.5em minus 0.4em\relax Springer Science \& Business Media, 2000.

\bibitem{poston2014catastrophe}
T.~Poston and I.~Stewart, \emph{Catastrophe theory and its applications}.\hskip 1em plus 0.5em minus 0.4em\relax Courier Corporation, 2014.

\bibitem{stulp2013robot}
F.~Stulp and O.~Sigaud, ``Robot skill learning: From reinforcement learning to evolution strategies,'' \emph{Paladyn, Journal of Behavioral Robotics}, vol.~4, no.~1, pp. 49--61, 2013.

\end{thebibliography}
\vspace{-1cm}
\begin{IEEEbiography}[{\includegraphics[width=1in,height=1.25in,clip,keepaspectratio]{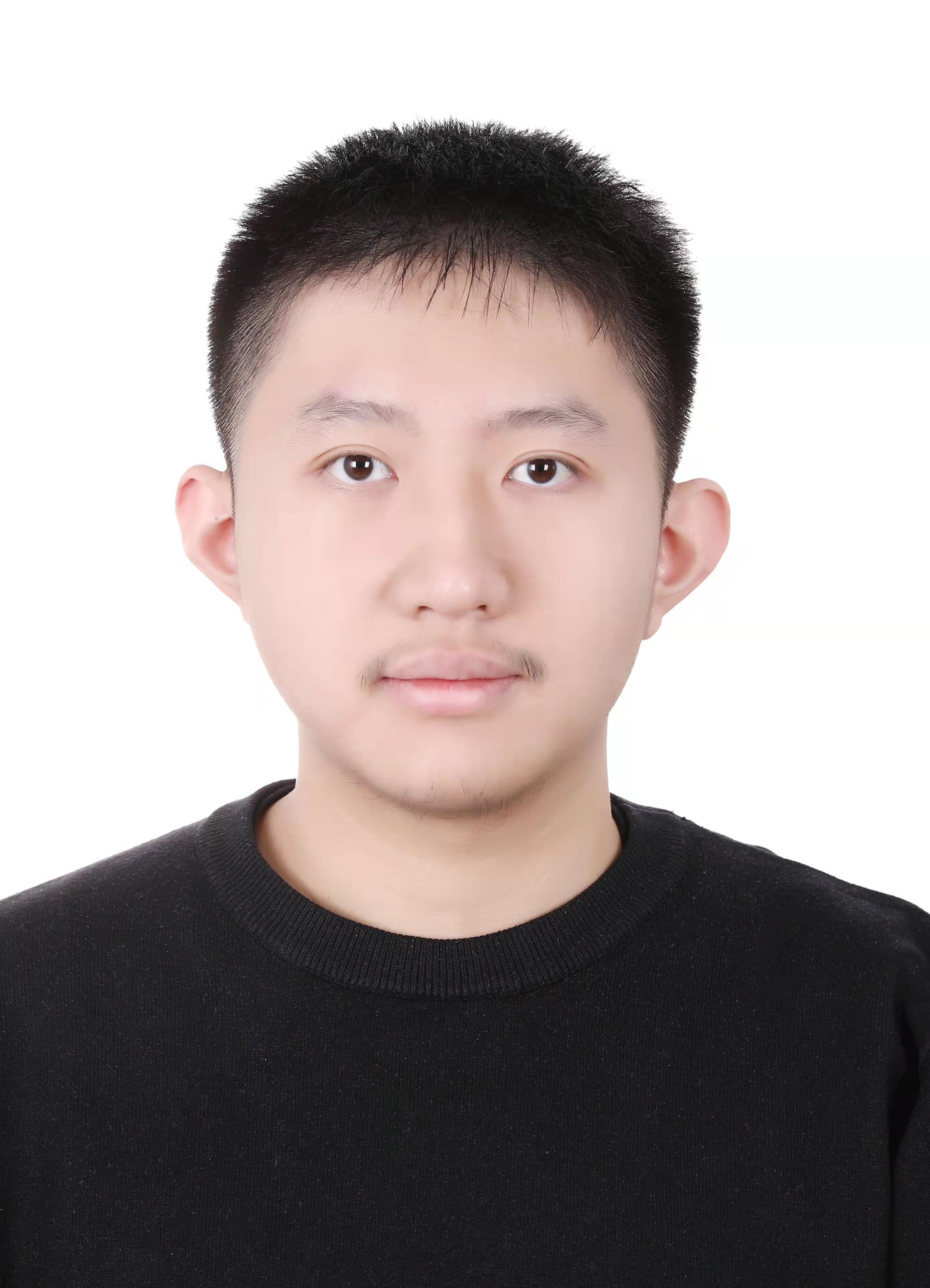}}]
{Lin Yang} received his Bachelor's degree from Beihang University, Beijing, China, in 2022. He is currently pursuing the Ph.D. degree under the supervision of Assoc. Prof. Lyu Chen and Assoc. Prof. Domenico Campolo from the school of MAE NTU. His current research interests include contact-rich manipulation via haptcs based SLAM, planning and sim2real.
\end{IEEEbiography}

\vspace{-1cm}
\begin{IEEEbiography}[{\includegraphics[width=1in,height=1.25in,clip,keepaspectratio]{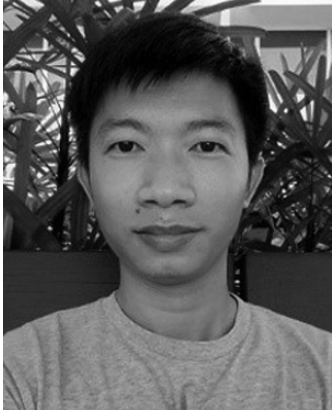}}]
{Huu-Thiet Nguyen} received the degree of engineer in control and automation engineering from Hanoi University of Science and Technology, Hanoi, Vietnam in 2015, and the PhD degree in electrical and electronic engineering from Nanyang Technological University, Singapore in 2022. He is currently a postdoctoral researcher at Nanyang Technological University. His research interests include robot control, robot learning, and machine learning in robotics and physical systems.
\end{IEEEbiography}

\vspace{-1cm}
\begin{IEEEbiography}[{\includegraphics[width=1in,height=1.25in,clip,keepaspectratio]{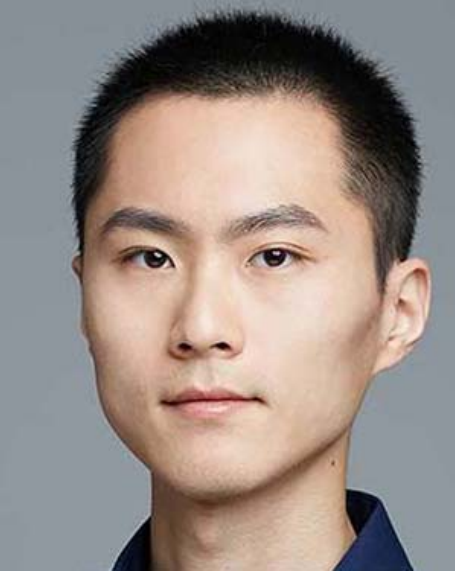}}]{Chen Lv}
(Senior Member, IEEE) received the Ph.D. degree from the Department of Automotive Engineering, Tsinghua University, China, in 2016. From 2014 to 2015, he was a Joint Ph.D. Researcher with the EECS Department, University of California at Berkeley. He is currently an Assistant Professor with Nanyang Technology University, Singapore. His research interests include cyber-physical systems, hybrid systems, advanced vehicle control, and intelligence, where he has contributed over 90 articles and holds 12 granted Chinese patents.
He received the Highly Commended Paper Award of IMechE, U.K., in 2012, the National Fellowship for Doctoral Student in 2013, the NSK Outstanding Mechanical Engineering Paper Award in 2014, China SAE Outstanding Paper Award in 2015, the 1st Class Award of China Automotive Industry Scientific and Technological Invention in 2015, Tsinghua University Outstanding Doctoral Thesis Award in 2016, and the IV2018 Best Workshop/Special Issue Paper Award. He serves as a Guest Editor for \textit{IEEE Intelligent Transportation Systems Magazine}, \textit{IEEE/ASME TRANSACTIONS ON MECHATRONICS}, and \textit{Applied Energy}; and an Associate Editor/Editorial Board Member for \textit{International Journal of Vehicle Autonomous Systems}, \textit{International Journal of Electric and Hybrid Vehicles}, and \textit{International Journal of Vehicle Systems Modelling and Testing}.
\end{IEEEbiography}

\vspace{-1cm}

\begin{IEEEbiography}[{\includegraphics[width=1in,height=1.25in,clip,keepaspectratio]{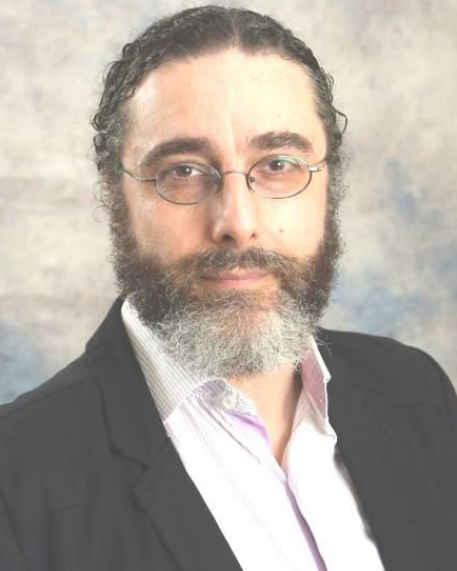}}]
{Domenico Campolo} received the Ph.D. degree in microengineering from Scuola Superiore Sant' Anna, Pisa, Italy, in 2002. He is currently
an Associate Professor and the Director of the Robotics Research Centre, School of Mechanical and Aerospace Engineering, Nanyang Technological University, Singapore. He is also the Co-Founder of ArtiCares Pte Ltd., an international company specializing in rehabilitation and
assistive robotics.
\end{IEEEbiography}

\end{document}